\documentclass[10pt,twocolumn,letterpaper]{article}

\usepackage{wacv}              

\usepackage{graphicx}
\usepackage{amsmath}
\usepackage{amssymb}
\usepackage{booktabs}
\usepackage{multirow}
\usepackage[ruled,vlined,linesnumbered]{algorithm2e}
\usepackage{xcolor}
\definecolor{commentGreen}{rgb}{0,0.5,0.05}

\definecolor{myRed}{rgb}{0.808,0.067,0.149}
\definecolor{myGreen}{rgb}{0.067,0.708,0.149}
\newcommand{\xmark}{{\color{myRed}\ding{55}}}%
\newcommand{\cmark}{{\color{myGreen}\ding{51}}} 
\usepackage{pifont} 
\usepackage{wrapfig}

\usepackage[pagebackref,breaklinks,colorlinks]{hyperref}

\usepackage[capitalize]{cleveref}
\crefname{section}{Sec.}{Secs.}
\Crefname{section}{Section}{Sections}
\Crefname{table}{Table}{Tables}
\crefname{table}{Tab.}{Tabs.}

\def\wacvPaperID{1373} 
\def\confName{WACV}
\def\confYear{2025}

\begin{document}

\title{Weight Copy and Low-Rank Adaptation for Few-Shot Distillation\\ of Vision Transformers\vspace{-0.3cm}}

\author{Diana-Nicoleta Grigore$^{1,\diamond}$, Mariana-Iuliana Georgescu$^{1,\diamond}$, Jon Alvarez Justo$^{2}$, Tor Johansen$^{2}$,\\
Andreea Iuliana Ionescu$^{3}$, Radu Tudor Ionescu$^{1,}$\thanks{Corresp.~author: raducu.ionescu@gmail.com.~~$^\diamond$Equal contribution.}\\
$^1$University of Bucharest, Romania, $^2$Norwegian University of Science and Technology, Norway,\\
$^3$University of Medicine and Pharmacy ``Carol Davila'', Romania\vspace{-0.3cm}}
\maketitle

\begin{abstract}
\vspace{-0.2cm}
Few-shot knowledge distillation recently emerged as a viable approach to harness the knowledge of large-scale pre-trained models, using limited data and computational resources. In this paper, we propose a novel few-shot feature distillation approach for vision transformers. Our approach is based on two key steps. Leveraging the fact that vision transformers have a consistent depth-wise structure, we first copy the weights from intermittent layers of existing pre-trained vision transformers (teachers) into shallower architectures (students), where the intermittence factor controls the complexity of the student transformer with respect to its teacher. Next, we employ an enhanced version of Low-Rank Adaptation (LoRA) to distill knowledge into the student in a few-shot scenario, aiming to recover the information processing carried out by the skipped teacher layers. We present comprehensive experiments with supervised and self-supervised transformers as teachers, on six data sets from various domains (natural, medical and satellite images) and tasks (classification and segmentation). The empirical results confirm the superiority of our approach over state-of-the-art competitors. Moreover, the ablation results demonstrate the usefulness of each component of the proposed pipeline. We release our code
at \small{\url{https://github.com/dianagrigore/WeCoLoRA}}.
\end{abstract}

\vspace{-0.3cm}
\section{Introduction}
\label{sec:intro}
\vspace{-0.1cm}

\begin{figure}[tb]
  \centering
  \includegraphics[width=1.0\linewidth]{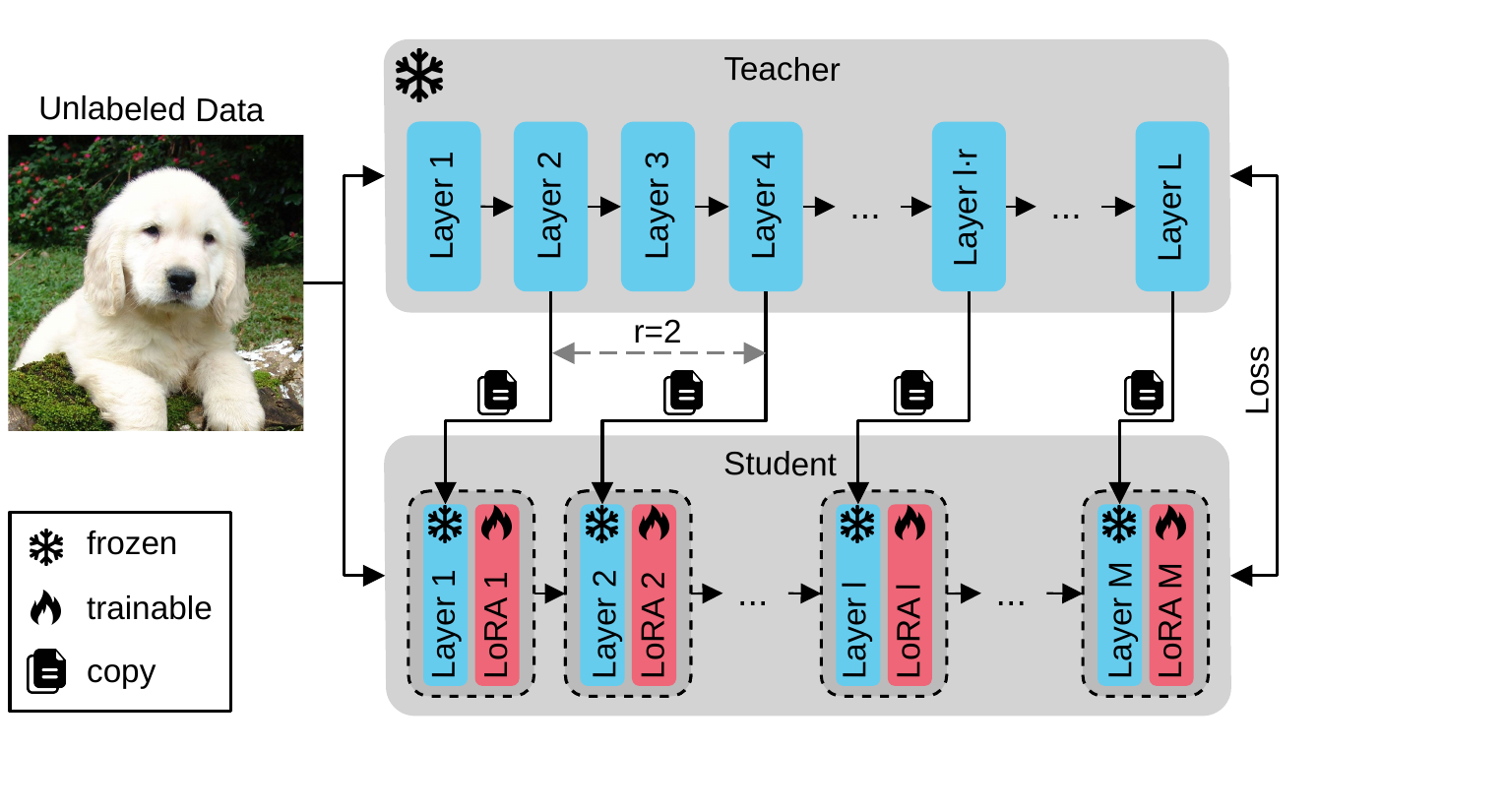}
      \vspace{-0.5cm}
  \caption{Our unsupervised feature distillation framework is based on two steps. In the first step, weights from intermittent layers of the teacher transformer are copied directly into the student, where the intermittence factor $r$ coincides with the compression ratio between the teacher and the student transformers. In the second step, enhanced LoRA blocks are integrated into the student network. The enhanced LoRA blocks are trained via feature distillation on unlabeled images. In the illustrated example, the compression ratio is $r=2$. Best viewed in color.} 
  \label{fig_pipeline}
      \vspace{-0.4cm}
\end{figure}

Vision transformers \cite{Bao-ICLR-2022,Dosovitskiy-ICLR-2021,Khan-CSUR-2022,Liu-ICCV-2021,Parmar-ICML-2018,Wu-ICCV-2021} have revolutionized computer vision research in the past few years, reaching state-of-the-art performance across a broad range of tasks, such as object recognition \cite{Dosovitskiy-ICLR-2021,Liu-ICCV-2021,Madan-WACV-2024,Wu-ICCV-2021}, object detection \cite{Carion-ECCV-2020,Li-ECCV-2022,Xie-NeurIPS-2021,Zheng-BMVC-2021,Zhu-ICLR-2020}, image segmentation \cite{Chen-arXiv-2021,Cheng-CVPR-2022,Gao-MICCAI-2021,Hatamizadeh-WACV-2022,Zhang-NeurIPS-2021}, image translation \cite{Kim-CVPR-2022,Ristea-NC-2023,Torbunov-WACV-2023}, among many others \cite{Khan-CSUR-2022}. Since the accuracy tends to grow as the model gets larger \cite{Dosovitskiy-ICLR-2021,Madan-WACV-2024}, most of the attention has been dedicated to building larger and more powerful models. However, the typically large size and slow inference speed of transformer-based architectures hinders the deployment of such models on environments with limited computational resources, \eg~on edge or mobile devices.

The success of vision transformers lies in the ``pre-training then fine-tuning'' paradigm, which originates from the natural language processing domain \cite{Vaswani-NIPS-2017,Devlin-NAACL-2019}. The multi-head attention layers inside transformers can capture long-range spatial relationships among tokens, but this flexibility has a significant downside: transformers can easily overfit small training sets and experience poor generalization capabilities \cite{Lee-Arxiv-2021,Liu-NeurIPS-2021}. Unlike convolutional nets, which benefit from the inductive bias of convolutional filters with small receptive fields \cite{Krizhevsky-CACM-2017}, transformers requires huge amounts of data in the pre-training stage to avoid overfitting \cite{Dosovitskiy-ICLR-2021}, regardless of the chosen supervised \cite{Dosovitskiy-ICLR-2021,Liu-ICCV-2021,Wu-ICCV-2021} or self-supervised \cite{Dong-AAAI-2023,Georgescu-ICCV-2023,He-CVPR-2022,Li-NeurIPS-2022,Tong-NeurIPS-2022,Wang-CVPR-2023} training setup. Therefore, to efficiently train large-scale transformer models, powerful and expensive computers are necessary, which are not widely available to researchers. Moreover, large amounts of data are not always available in some domains, \eg~hyperspectral image segmentation \cite{Justo-Arxiv-2023}.

To mitigate the challenges of training large-scale models on large amounts of data on machines with limited computational resources, researchers have proposed the few-shot knowledge distillation (FSKD) paradigm \cite{He-NN-2022,Li-CVPR-2020,Nguyen-ECCV-2022,Pan-ACL-2021,Rajasegaran-BMVC-2021,Sauer-NLP4ConvAI-2022,Shen-AAAI-2021,Zhang-ECAI-2020,Zhou-ACL-2022}, which was explored in both language \cite{Pan-ACL-2021,Sauer-NLP4ConvAI-2022,Zhou-ACL-2022} and vision \cite{He-NN-2022,Li-CVPR-2020,Lim-NC-2021,Lin-CVRP-2023,Nguyen-ECCV-2022,Shen-AAAI-2021,Zhang-ECAI-2020} domains. This paradigm allows the trained model (called student) to benefit from the knowledge learned by large-scale transformers (called teachers), while significantly reducing the training time and data set size. 

We emphasize that FSKD has not been extensively explored in the vision domain \cite{He-NN-2022,Li-CVPR-2020,Lim-NC-2021,Lin-CVRP-2023,Nguyen-ECCV-2022,Rajasegaran-BMVC-2021,Shen-AAAI-2021,Zhang-ECAI-2020}, with even less studies focused on the pre-training stage of vision transformers \cite{Lin-CVRP-2023}. To this end, we propose a novel few-shot feature distillation approach for vision transformers based on intermittent \textbf{We}ight \textbf{Co}pying and \textbf{Lo}w-\textbf{R}ank \textbf{A}daptation (WeCoLoRA), as illustrated in Figure \ref{fig_pipeline}. Our approach is divided into two steps. For the first step, we leverage the fact that vision transformers have a consistent depth-wise structure, \ie the input and output dimensions are compatible across transformer blocks. This allows us to directly copy the weights from intermittent layers of existing pre-trained vision transformers (teachers) into shallower architectures (students). Here, we use the intermittence factor to control the complexity (size) of the student transformer with respect to its teacher. In the second step, we employ an enhanced version of Low-Rank Adaptation (LoRA) \cite{Hu-ICLR-2022} to distill knowledge into the student in a few-shot scenario, aiming to recover the information processing carried out by the teacher layers that were skipped in the first step. We hereby note that standard LoRA is usually employed over queries and values, which is insufficient to fully replicate the behavior of the skipped blocks. To mitigate this issue and better complement the weight copying mechanism, we apply enhanced LoRA, a variant of LoRA that approximates all layers in the skipped transformer block.

We perform the pre-training stage of efficient student transformers via few-shot knowledge distillation on various subsets of ImageNet-1K \cite{Deng-CVPR-2009}. Then, we carry out linear probing experiments on five downstream data sets from different image domains, namely ImageNet-1K \cite{Deng-CVPR-2009}, CIFAR-100 \cite{Krizhevsky-TR-2009}, ChestX-ray14 \cite{Wang-CVPR-2017}, iNaturalist \cite{VanHorn-CVPR-2018} and RESISC45 \cite{Cheng-PIEEE-2017}. We compare WeCoLoRA with state-of-the-art competitors, namely DeiT \cite{Touvron-ICML-2021}, DMAE \cite{bai2023masked}, MiniViT \cite{Zhang-CVPR-2022} and DoRA \cite{Liu-ICML-2024}, showing that our approach leads to superior results. Furthermore, we conduct an ablation study to demonstrate that each proposed component brings significant performance gains. Additionally, we perform an analysis of the distilled features, which explains why WeCoLoRA produces more robust features than competing methods. In the supplementary, we demonstrate the versatility of WeCoLoRA by applying it on a convolutional network for image segmentation on the MSLesSeg \cite{Rondinella-CBM-2023} data set.

In summary, our contribution is threefold:
\begin{itemize}
    \item \vspace{-0.15cm} We propose a novel unsupervised few-shot feature distillation approach for vision transformers based on $(i)$ intermittent weight copying and $(ii)$ enhanced low-rank adaption, called WeCoLoRA.
    \item \vspace{-0.18cm} We present few-shot and linear probing experiments on six benchmark data sets comprising natural, medical and satellite images, demonstrating the utility of our training pipeline across different domains and tasks.
    \item \vspace{-0.18cm} We analyze the features learned by our distilled models in comparison with those of the strongest competitor, showing that our approach generates more robust and discriminative features. 
\end{itemize}

\section{Related Work}
\label{sec:related_work}  

\noindent
\textbf{Knowledge distillation.} Knowledge distillation (KD) \cite{hinton2015distilling, lopez2015unifying, romero2014fitnets, mirzadeh2020improved}, a.k.a.~teacher-student training, is an efficiency boosting technique meant to reduce the computational load that comes with large models. It emerged \cite{lopez2015unifying} from the union of model compression \cite{hinton2015distilling, ba2014deep} and learning under privileged information \cite{vapnik2009new}. Our study is only preoccupied with the former task, being aimed at proposing an approach that involves transferring expertise from a heavy teacher model to a lighter student model \cite{bucilua2006model, urban2016deep}, with the latter finally being able to learn more discriminative features than through a conventional training procedure \cite{Cho_2019_ICCV, ji2020knowledge, cheng2020explaining}. 

\noindent
\textbf{KD for vision transformers.} KD was originally applied to vision transformers by Touvron \etal~\cite{Touvron-ICML-2021}, who proposed Data-Efficient Image Transformers (DeiT) based on leveraging attention as knowledge and enhancing the student's capabilities through a learned distillation token. Since then, several other studies used KD on vision transformers \cite{Lin-CVRP-2023, kelenyi2024sam, zhao2023cumulative, hao2022learning, wu2022tinyvit}. Close to our work, MiniViT \cite{Zhang-CVPR-2022} uses a parameter reduction technique that involves copying weights (and applying slight perturbations) between the teacher and a smaller student transformer, followed by distillation through self-attention. In comparison, after copying the weights, we employ an enhanced version of LoRA \cite{Hu-ICLR-2022} during the information recovery process, which leads to significantly better results in the few-shot scenario. Recently, distilling Masked Autoencoders \cite{He-CVPR-2022} has gathered some traction \cite{Lin-CVRP-2023, bai2023masked, lao2023masked}, with most approaches achieving data efficiency through masking very large percentages (up to 98\%) of the input. Our data reduction strategy comes solely from a reduction of the number of unlabeled samples, being more fit for cases where data scarcity is an issue.

\noindent
\textbf{Few-shot knowledge distillation.} Zero-shot knowledge distillation \cite{Barbalau-NeurIPS-2020,zhu2021data, lopes2017data, zhang2022fine, chawla2021data} implies that the teacher and student do not share the same data sources. As a consequence, the latter should obtain its knowledge through synthetic examples \cite{chen2019data, liu2018model}. Black-box KD \cite{Barbalau-NeurIPS-2020,Nguyen-ECCV-2022} involves generating synthetic images from the teacher's training distribution, along with their corresponding labels, offering the dual benefit of privacy preservation and effective learning. Zero-shot KD emerged as a solution to the lack of knowledge about the training data used by the teacher, but it suffers from the problem of collapse \cite{do2022momentum}, caused by similar training samples. A less strict alternative that mitigates this issue is to employ few-shot KD \cite{He-NN-2022,Li-CVPR-2020,Lim-NC-2021,Lin-CVRP-2023,Nguyen-ECCV-2022,Rajasegaran-BMVC-2021,Shen-AAAI-2021,Zhang-ECAI-2020}. To perform FSKD, Li \etal~\cite{Li-CVPR-2020} introduced an additional $1\times1$ convolutional layer at the end of each block of a CNN to support recovering the abilities of a teacher, using a small set of unlabeled examples. To the best of our knowledge, we are the first to employ few-shot KD in the pre-training stage of lightweight vision transformers.

\noindent
\textbf{Low-Rank Adaptation.} LoRA and its variations \cite{Hu-ICLR-2022, Liu-ICML-2024, zhu2023sira, xu2023qa} represent parallel performance improvement methods, enabling the use of powerful large models through lightweight fine-tuning. Consequently, LoRA reduces the inference cost and the requirement to use large data sets. Following other lines of work, the authors of LoRA \cite{Hu-ICLR-2022} showed that over-parameterized models rely on a very low internal dimension when making their decisions \cite{oymak2019generalization}. The proven hypothesis is that the change in weights during domain adaptation or fine-tuning should also be happening at a low intrinsic rank. To boost training efficiency, we thus combine Low-Rank Adaptation \cite{Hu-ICLR-2022} and KD \cite{hinton2015distilling} in our study. To the best of our knowledge, there is no prior work that employs weight copying and LoRA as a method for few-shot feature distillation in an unsupervised setting.

\section{Method}
\label{sec:method} 
\vspace{-0.1cm}
\subsection{WeCoLoRA Architecture} 
\label{sec:WeCoLoRA} 
\vspace{-0.1cm}

Our knowledge distillation framework, WeCoLoRA, is formally described in Algorithm~\ref{alg:LORA_KD}. Our method follows the knowledge distillation paradigm of compressing a larger model into a smaller one. Therefore, our goal is to obtain a student \textbf{S} which runs faster than the teacher \textbf{T} during inference. However, we also use significantly less training data during the knowledge distillation process, since it is not always feasible to obtain the same amount of data as for (pre-)training the teacher. Our knowledge distillation method is designed for vision transformers~\cite{Dosovitskiy-ICLR-2021}. 
Given a pre-trained transformer-based teacher \textbf{T} with $L$ layers, we create the student \textbf{S} that has $\lfloor\frac{L}{r}\rfloor$ layers, where $r$ is the reduction ratio of the number of layers and $\lfloor\cdot\rfloor$ is the floor approximation function.
The first step of our algorithm, which is described in steps $2$-$4$ of Algorithm~\ref{alg:LORA_KD}, is to copy the pre-trained weights of \textbf{T} to \textbf{S}. Since the student \textbf{S} has fewer layers than the teacher, due to the reduction ratio $r$, the weights from the layer at index $r\cdot l$ of the teacher are copied to the layer at index $l$ of the student, where $l \in \{1, 2,..., \lfloor\frac{L}{r}\rfloor\}$.

After the intermittent weight copying step, we apply an adaptor layer, such that the student model is able to recover the weights which are not transferred from the teacher. In this way, we aim to minimize the performance gap between the teacher and the student.
We employ the LoRA framework \cite{luo2023lcm} as the adaptor, due to its competitive performance obtained in model adaptation.
LoRA can be applied to any fully-connected layer, which makes it a great choice for the transformer architecture.
Therefore, for a pre-trained fully-connected layer $f$, with the weight matrix $\boldsymbol{W} \in \mathbb{R}^{d_1 \times d_2}$, LoRA learns a matrix $\boldsymbol{B} \in \mathbb{R}^{d_1 \times k}$ and a matrix $\boldsymbol{A}  \in \mathbb{R}^{k \times d_2}$, where $k$ is the matrix rank, $k \leq min(d_1, d_2)$, and $d_1$ and $d_2$ are the matrix dimensions.
During the knowledge distillation process, $\boldsymbol{W}$ is kept frozen, so only the parameters $\boldsymbol{A}$ and $\boldsymbol{B}$ are updated. The output of the layer $f$ becomes $f(x)=\boldsymbol{W}\cdot x + \boldsymbol{B}\cdot\boldsymbol{A}\cdot x$, where $x \in \mathbb{R}^{d_1}$ is the input and $f(x) \in \mathbb{R}^{d_1\times d_2}$.

Luo \etal~\cite{luo2023lcm} proposed to apply LoRA only to the projection matrices corresponding to the query and value tokens $\boldsymbol{Q}$ and $\boldsymbol{V}$, respectively. However, in our setup, after applying the intermittent weight copying mechanism, we need to replicate the functionality of the skipped transformer blocks. Since LoRA only adapts the projection matrices of query and value tokens, it does not provide the means to fully replicate the skipped blocks. In order to mitigate this issue, we conjecture that it is better to apply LoRA to each component of the transformer block. Therefore, for each head $h_i$, $\forall i \in \{1,2,...,H\}$ of the multi-head self-attention (MSA) layer, the outputs of the query $\boldsymbol{Q}_i$, value $\boldsymbol{V}_{\!i}$ and key $\boldsymbol{K}_i$ applied to the input $x$ become:
\begin{equation}\label{new_qvk}
\begin{split}
    \boldsymbol{Q}_i &= \boldsymbol{W}_{\!Q_i}\cdot x + \boldsymbol{B}_{Q_i}\cdot \boldsymbol{A}_{Q_i} \cdot x, \\
    \boldsymbol{K}_i &= \boldsymbol{W}_{\!K_i}\cdot x + \boldsymbol{B}_{K_i}\cdot \boldsymbol{A}_{K_i}\cdot x, \\
    \boldsymbol{V}_i &= \boldsymbol{W}_{\!V_i}\cdot x + \boldsymbol{B}_{V_i}\cdot\boldsymbol{A}_{V_i}\cdot x,
\end{split} 
\end{equation}
where $\boldsymbol{W}_{\!Q_i}$, $\boldsymbol{W}_{\!K_i}$ and $\boldsymbol{W}_{\!V_i}$ are the query, key and value projection matrices, while $\boldsymbol{B}_{*_i}$ and $\boldsymbol{A}_{*_i}$ are the matrices learned by LoRA, where $*\in \{ Q, K, V \}$.

The formula to compute the output of the head $h_i$ remains unaltered, being equal to $h_i=\mbox{Attention}(\boldsymbol{Q}_i, \boldsymbol{K}_i, \boldsymbol{V}_{\!i})$, $\forall i \in \{1,2,...,H\}$, where:
\begin{equation}\label{head} 
    \mbox{Attention}(\boldsymbol{Q}, \boldsymbol{K}, \boldsymbol{V}) = \mbox{softmax}\!\left( \frac{\boldsymbol{Q}\cdot \boldsymbol{K}^{\top}}{\sqrt{d_k}} \right) \cdot \boldsymbol{V},
\end{equation} and $d_k$ is the dimension of $\boldsymbol{K}$.

\begin{algorithm*}[!t]
\caption{Knowledge Distillation with WeCoLoRA}
\label{alg:LORA_KD} 
 
\KwIn{$\mathcal{D}=\{ \boldsymbol{I}_1, \boldsymbol{I}_2, .., \boldsymbol{I}_n \}$ - the training set of unlabeled images, \textbf{T} - the pre-trained teacher model, $\theta_{\textbf{T}}$ - pre-trained weights of the teacher, $\eta$ - the learning rate, $r$ - the depth reduction factor, $k$ - the rank of the low-rank matrices, $L$ - the number of layers of the teacher \textbf{T}.}


\KwOut{
$\theta_\textbf{S}$ - the trained weights of the student model.
}
$n \gets \lvert \mathcal{D} \rvert;$ \textcolor{commentGreen}{$\lhd$ get the number of training samples}\\  
$M \gets \lfloor \frac{L}{r} \rfloor;$ \textcolor{commentGreen}{$\lhd$ compute the number of layers of the student \textbf{S}}\\
\ForEach{$l \in \{1,2,...,M\}$} 
{
   $\theta_{\textbf{S}_l} \gets \theta_{\textbf{T}_{l\cdot r}};$  \textcolor{commentGreen}{$\lhd$ copy weights from the teacher} \\
}

$\theta^+_\textbf{S}  \gets \emptyset;$  \textcolor{commentGreen}{$\lhd$ initialize trainable weights} \\

\ForEach{$l \in \{1,2,...,M\}$} 
{
   $\textbf{S}_l \gets \textbf{S}_l \cup \text{WeCoLoRA}(\textbf{S}_l);$  \textcolor{commentGreen}{$\lhd$ add adaptor, as described in Section~\ref{sec:WeCoLoRA}} \\
    $\theta^+_{\textbf{S}_l} \gets \theta^+_{\textbf{S}_l} \cup \{ \boldsymbol{A}_{*_l}, \boldsymbol{B}_{*_l} \};$ \textcolor{commentGreen}{$\lhd$ add the new weights}
}

\Repeat{convergence}{
        \ForEach{$i \in \{1,2,...,n\}$}{
            $\boldsymbol{E}^\textbf{T}_i \gets \textbf{T}(\boldsymbol{I}_i,\theta_\textbf{T});$ \textcolor{commentGreen}{$\lhd$ get embedding from teacher}\\
            $\boldsymbol{E}^\textbf{S}_i \gets \textbf{S}(\boldsymbol{I}_i, \theta_\textbf{S}  \cup \theta^+_\textbf{S} );$ \textcolor{commentGreen}{$\lhd$ get embedding from student}\\
            $\mathcal{L}_{\scriptsize{\mbox{KD}}} \gets \mathcal{L}(\boldsymbol{E}^\textbf{T}_i, \boldsymbol{E}^\textbf{S}_i);$ \textcolor{commentGreen}{$\lhd$ apply Eq.~\eqref{kd}} \\
            $\theta^+_\textbf{S} \gets \theta^+_\textbf{S} - \eta \cdot \nabla \mathcal{L}_{\scriptsize{\mbox{KD}}};$ \textcolor{commentGreen}{$\lhd$ update only the newly added weights} \\
        }
    }

$\theta_\textbf{S} \gets \theta_\textbf{S} \oplus \theta^+_\textbf{S};$  \textcolor{commentGreen}{$\lhd$ integrate trained weights into the copied weights} \\
\end{algorithm*}

We also add adaptation parameters to the output projection layer in the multi-head attention, resulting in:
\begin{equation} 
\begin{split}
   \mbox{MultiHead}(\boldsymbol{Q}, \boldsymbol{K}, \boldsymbol{V}) &= \boldsymbol{W}_{\!O}\cdot[h_1, h_2, .., h_H] + \\
   &+  \boldsymbol{B}_{O}\cdot \boldsymbol{A}_{O}\cdot [h_1, h_2, .., h_H],
\end{split}
\end{equation}
where [$\cdot$] is the concatenation operation, $\boldsymbol{W}_{\!O}$ is the output projection matrix, $\boldsymbol{B}_{O}$ and $\boldsymbol{A}_{O}$ are the parameters introduced by LoRA, and $H$ is the number of heads. 

Further, we also integrate LoRA into the feed-forward networks (FFNs). Therefore, the output of an FFN from a transformer block is:
\begin{equation}\label{ffn} 
\begin{split}
        \text{FFN}(x) &= \boldsymbol{W}_{\!f_2} \cdot \text{FFL}(x) +  \boldsymbol{B}_{f_2}\cdot\boldsymbol{A}_{f_2}\cdot \text{FFL}(x), \\
        \text{FFL}(x) &= \sigma(\boldsymbol{W}_{\!f_1} \cdot x + \boldsymbol{B}_{f_1}\cdot\boldsymbol{A}_{f_1}\cdot x),
\end{split}
\end{equation}
where FFL is the output of the first layer in the FFN model, $\boldsymbol{W}_{\!f_1}$ and $\boldsymbol{W}_{\!f_2}$ are the parameters of the feed-forward network, $\sigma$ is the activation function, and $\boldsymbol{A}_{*}$ and $\boldsymbol{B}_{*}$ are the matrices learned by LoRA, where $*\in \{f_1, f_2\}$. The biases in the FFN network were omitted to enhance the clarity of the presentation.
By introducing LoRA in every component of the transformer block, we obtain an enhanced version of LoRA-based transformer blocks. Enhanced LoRA is able to better complement the weight copying mechanism by enabling a more faithful replication of the skipped teacher blocks. This statement is confirmed by our ablation study.

\vspace{-0.1cm}
\subsection{Knowledge Distillation with WeCoLoRA} 
\vspace{-0.1cm}

Our distillation method does not require training labels, therefore, the knowledge distillation data set is $\mathcal{D}=\{ \boldsymbol{I}_1, \boldsymbol{I}_2, .., \boldsymbol{I}_n \}$, where $\boldsymbol{I}_i$ is an image sample and $n$ is the number of samples in the data set. This is because the knowledge distillation procedure is performed in the latent space. Moreover, we chose $n$ to be typically small, resulting in a few-shot feature distillation training setup.

The goal of applying knowledge distillation is to transfer the knowledge of the pre-trained teacher \textbf{T} into the LoRA-enhanced student \textbf{S}.
Following Luo \etal~\cite{luo2023lcm}, during optimization, we only update the newly added weights of the student \textbf{S}, denoted as $\theta^+_\textbf{S}=\{\boldsymbol{A}_*, \boldsymbol{B}_*\}$, where $* \in \{Q_l, K_l, V_l, O_l, f_{l,1}, f_{l,2}\}$, $\forall l \in \{1, 2,..., \lfloor\frac{L}{r}\rfloor\}$.  The operation of inserting trainable weights into the student is formally described in steps $6$-$8$ of Algorithm~\ref{alg:LORA_KD}. 

In order to optimize the parameters $\theta_\textbf{{S}}^+$ of the student, we minimize the absolute difference between the embedding $\boldsymbol{E}^{\textbf{T}}_i \in \mathbb{R}^{t \times d}$ returned by the teacher \textbf{T} for image $\boldsymbol{I}_i$, and the embedding $\boldsymbol{E}^{\textbf{S}}_i \in \mathbb{R}^{t \times d}$ returned by the student \textbf{S} for the same image $\boldsymbol{I}_i$, where $t$ is the number of tokens and $d$ is their hidden dimension. Formally, the proposed feature distillation is expressed as follows:
\begin{equation}\label{kd} 
\begin{split}
\mathcal{L}(\boldsymbol{E}^{\textbf{T}}_i, \boldsymbol{E}^{\textbf{S}}_i) = \left| \boldsymbol{E}^{\textbf{T}}_i - \boldsymbol{E}^{\textbf{S}}_i \right|, \forall i \in \{1,2,...,n\},
\end{split}
\end{equation}
where $n$ is the number of training images. This procedure is described in steps $10$-$14$ of Algorithm~\ref{alg:LORA_KD}. 

After optimizing the parameters $\theta_\textbf{\text{S}}^+$, we integrate them into the pre-trained  parameters $\theta_\textbf{{S}}$ (copied from the teacher \textbf{T}) in step 16, as described by Luo \etal~\cite{luo2023lcm}, in order to avoid affecting the running time during inference.

\section{Experiments}
\vspace{-0.1cm}

We use ImageNet-1K \cite{Deng-CVPR-2009} to perform knowledge distillation in various few-shot setups, considering subsets ranging between $0.25\%$ and $10\%$ of the number of samples in the original training set. We test the models via linear probing, using the official ImageNet-1K evaluation set. We also perform linear probing experiments on five downstream data sets: ImageNet-1K \cite{Deng-CVPR-2009}, CIFAR-100~\cite{Krizhevsky-TR-2009}, ChestX-ray14~\cite{Wang-CVPR-2017}, iNaturalist~\cite{VanHorn-CVPR-2018} and RESISC45~\cite{Cheng-PIEEE-2017}.

\vspace{-0.1cm}
\subsection{Data Sets}
\vspace{-0.1cm}

\noindent
\textbf{ImageNet-1K}. In our self-supervised knowledge distillation process, we operate on ImageNet-1K \cite{Deng-CVPR-2009}, a large-scale data set with 1,281,167 color images from various sources, each having an average shape of $469\times387$ pixels. The collection covers 1,000 object categories, such as plants, vehicles, and everyday items.

\noindent
\textbf{CIFAR-100}. CIFAR-100 \cite{Krizhevsky-TR-2009} is a balanced data set of 60,000 images representing various visual entities, grouped into 100 classes. The categories are arranged in a hierarchical structure, under 20 superclasses. The official test set contains 10,000 pictures. Each image has $32\times 32$ pixels.

\noindent
\textbf{ChestX-ray14}. The CXR14 \cite{Wang-CVPR-2017} data set is composed of 112,120 frontal-view X-Ray images from 30,805 patients and depicts a collection of 14 common pathologies, as well as normal lung scans. Its annotations are binary multi-labels, indicating whether each of 14 lung diseases is present or not. The labels are based on radiological reports. The resolution of each image is $1024\times 1024$. To evaluate our models, we use the test set comprising 20\% of the images, on which we perform multi-label classification. 

\noindent
\textbf{iNaturalist}. iNaturalist \cite{VanHorn-CVPR-2018} is a heavily imbalanced data set, proposing the fine-grained classification of 1,010 plant and animal species. Comprising a total size of 268,243 images with varying resolutions, its 2019 version offers an evaluation set of 3,030 images. 

\noindent
\textbf{RESISC45}. NWPU-RESISC45 \cite{Cheng-PIEEE-2017} incorporates 31,500 3-channel images from the aerial domain, which can be used for remote sensing classification. The data set comprises 45 scene classes and various spatial resolutions, where one pixel can represent a surface ranging from 20cm to 30m. In our evaluation setting, we use 80\% of the data as the test set. 

\vspace{-0.1cm}
\subsection{Implementation Details}
\vspace{-0.1cm}

We implement our method in Pytorch. We create the enhanced version of LoRA starting from the official code of LoRA\footnote{\url{https://github.com/microsoft/LoRA}} and extending it, in order to be applied to each component of the transformer layer. To demonstrate the generalization of our method to different teachers, we perform experiments with a supervised and a self-supervised teacher, respectively. As the supervised teacher backbone, we employ the ViT-B model \cite{Dosovitskiy-ICLR-2021} trained on ImageNet, available in the \texttt{timm}\footnote{\url{https://pypi.org/project/timm}} library. The selected self-supervised model is also based on ViT-B, but pre-trained using the Masked Autoencoder framework~\cite{He-CVPR-2022}. Since the goal is to create a model which is faster and lighter, we generally refrain from using larger teacher models, which would be inefficient for our purpose.

To show that our method generalizes to different depth reduction factors, we perform experiments with $r\!=\!2$, $r\!=\!3$ and $r\!=\!4$. Since the number of layers $L$ in the teacher architecture (ViT-B) is $12$, the number of layers in the student architecture becomes $M\!=\!6$ (when $r\!=\!2$), $M\!=\!4$ (when $r\!=\!3$), or $M\!=\!3$ (when $r\!=\!4$), respectively.
We select the rank of the low-rank matrices from $k \in \{128, 256, 512\}$, which controls the number of parameters used to recover the skipped layers from the teacher. 

We perform the knowledge distillation procedure for only $10$ epochs. We set the learning rate to $10^{-3}$ when the compression factor $r$ equals $2$, and to $10^{-4}$ when $r$ is larger. The batch size is set to $128$, with a gradient accumulation factor of $8$. We perform weak data augmentation, similar to He \etal~\cite{He-CVPR-2022}.

To demonstrate that our method transfers strong features, we only perform linear probing on the downstream tasks. For each downstream data set, we train the linear classifier using the best available linear probing training recipe. We also provide the code for an easier reproduction of the experiments. For a fair comparison, we use the same linear probing training setting for all the models included in the comparison.

\vspace{-0.1cm}
\subsection{Results}
\label{sec:ablation} 
\vspace{-0.1cm}

\begin{table}[!t]
    \centering
    \setlength\tabcolsep{2.0pt}
    \small{
    \begin{tabular}{|c|c|c|c|c|c|c|}
        \hline
        $\alpha$     &   \rotatebox[origin=c]{60}{Weight Copy}  & \rotatebox[origin=c]{60}{$\;\;\;\;$LoRA$\;\;\;\;$}  & \rotatebox[origin=c]{60}{$\;$Enhanced LoRA$\;$} &    \rotatebox[origin=c]{60}{ImageNet} & \rotatebox[origin=c]{60}{iNaturalist} & \rotatebox[origin=c]{60}{CIFAR-100} \\
        \hline
        \hline
       \multirow{4}{*}{\rotatebox{90}{$\alpha=1\%$}} & \xmark & \xmark &  \xmark &  $3.5$  &  $2.4$   &  $9.1$ \\
                                   & \cmark & \xmark &  \xmark &   $30.2$ & $19.2$ & $32.1$\\
                                   & \cmark &  \cmark &  \xmark   &  $32.4$  &  $21.6$   &  $30.7$\\ 
                                   & \cmark & \xmark &  \cmark  & $\mathbf{36.8}$ & $\mathbf{22.7}$ & $\mathbf{33.3}$\\
                              
       \hline                
        \multirow{4}{*}{\rotatebox{90}{$\alpha=10\%$}} & \xmark & \xmark &  \xmark &  $15.8$  &  $14.5$   &  $28.9$  \\
                                   & \cmark & \xmark &  \xmark & $53.7$  & $31.2$ & $48.2$ \\
                                   & \cmark &  \cmark &  \xmark   & $33.0$  & $28.4$  & $41.1$ \\
                                   & \cmark & \xmark &  \cmark  & $\mathbf{56.0}$ & $\mathbf{34.5}$  & $\mathbf{49.8}$ \\
                                   
        \hline 
        
    \end{tabular}
    }
        \vspace{-0.3cm}
        \caption{Ablation results in terms of accuracy (in percentages) on ImageNet-1K~\cite{Deng-CVPR-2009}, iNaturalist19~\cite{VanHorn-CVPR-2018} and CIFAR-100~\cite{Krizhevsky-TR-2009}. The results are obtained using a compression factor of $r=2$, employing the self-supervised teacher ViT-B~\cite{He-CVPR-2022}. The parameter $\alpha$ represents the percentage of the original ImageNet-1K training set~\cite{Deng-CVPR-2009} used during knowledge distillation. The best accuracy on each downstream data set and each value of $\alpha$ is highlighted in bold.}
    \label{tab:comp_to_kd}
        \vspace{-0.2cm}
\end{table} 

\begin{table*}[t]
    \centering
    \setlength\tabcolsep{2.5pt}
    \small{
    \begin{tabular}{|c|l|c|c|c|c|c|c|c|}
        \hline
        \multirow{2}{*}{$\alpha$}     &   \multirow{2}{*}{Method} & \multirow{2}{*}{Venue} & \multicolumn{3}{c|}{Self-supervised teacher} & \multicolumn{3}{c|}{Supervised teacher}\\
        \cline{4-9}
        & &        &  {ImageNet} & {iNaturalist} & {CIFAR-100} &  {ImageNet} & {iNaturalist} & {CIFAR-100} \\
        \hline
        \hline
       \multirow{6}{*}{$\alpha=1$}  
            & DeiT \cite{Touvron-ICML-2021}  & ICML 2021 & $16.5$ & $12.9$  & $28.8$ & $25.5$ & $20.3$ & $39.0$\\
            & MiniViT \cite{Zhang-CVPR-2022} & CVPR 2022 & - & - & - & $26.7$ & $21.0$ & $37.3$ \\
            & DMAE \cite{bai2023masked}  & CVPR 2023 &  $15.5$ & $8.5$ & $22.4$ & $15.4$ & $7.9$ & $22.2$ \\
            & DoRA \cite{Liu-ICML-2024} & ICML 2024 & - & - & - & $29.0$ & $20.9$ & $25.6$ \\
            
            & WeCo+KD (ours, ablated) & - &   $30.2$ & $19.2$ & $32.1$ & $35.4$ & $25.4$ & $34.8$ \\
            & WeCoLoRA (ours) & - & $\mathbf{36.8}$ & $\mathbf{22.7}$ & $\mathbf{33.3}$ & $\mathbf{60.5}$ & $\mathbf{43.4}$ & $\mathbf{60.6}$  \\
       \hline                
        \multirow{6}{*}{$\alpha=10$}  
                                  
            & DeiT \cite{Touvron-ICML-2021} & ICML 2021 & $39.9$ & $29.2$  & $48.1$ & $57.6$ & $37.8$ & $61.4$ \\
            & MiniViT \cite{Zhang-CVPR-2022} & CVPR 2022 & - & - & - & $64.0$ & $40.8$ & $64.2$ \\
            & DMAE \cite{bai2023masked} & CVPR 2023 & $30.0$  &  $17.1$  & $31.7$ & $33.0$ & $21.0$ & $35.6$ \\
            & DoRA \cite{Liu-ICML-2024} & ICML 2024 & - & - & - & $58.7$ & $45.3$ & $58.7$ \\

            & WeCo+KD (ours, ablated) & - & $53.7$  & $31.2$ & $48.2$ & $68.3$ & $45.7$ & $64.4$ \\
            & WeCoLoRA (ours) & - &  $\mathbf{56.0}$ & $\mathbf{34.5}$  & $\mathbf{49.8}$ & $\mathbf{69.2}$ & $\mathbf{49.5}$ & $\mathbf{68.3}$ \\
        \hline  
          $\alpha=100$ & ViT-B (teacher)  & -         & $66.1$ & $41.6$  & $55.1$ & $84.2$ & $52.5$ & $89.1$ \\
          \hline
    \end{tabular}
    }
        \vspace{-0.3cm}
        \caption{Accuracy rates (in percentages) obtained on ImageNet-1K~\cite{Deng-CVPR-2009}, iNaturalist19~\cite{VanHorn-CVPR-2018} and CIFAR-100~\cite{Krizhevsky-TR-2009} by four state-of-the-art methods, DeiT \cite{Touvron-ICML-2021}, MiniViT \cite{Zhang-CVPR-2022}, DMAE \cite{bai2023masked} and DoRA \cite{Liu-ICML-2024}, in comparison with WeCo+KD and WeCoLoRA. The results are obtained using a compression factor of $r=2$, while alternating between the self-supervised ViT-B teacher~\cite{He-CVPR-2022} and the supervised ViT-B teacher \cite{Dosovitskiy-ICLR-2021}. The parameter $\alpha$ represents the percentage of the original ImageNet-1K training set~\cite{Deng-CVPR-2009} used during knowledge distillation. The best accuracy on each downstream data set and each value of $\alpha$ is highlighted in bold. The teachers are based on linear probing.}
    \label{tab:comp_to_sota_kd}
    \vspace{-0.1cm}
\end{table*} 

\noindent
\textbf{Ablation study.} 
We compare our method with various ablated versions and report the results in Table~\ref{tab:comp_to_kd}. We start by comparing our method to the conventional KD method based on training a student vision transformer from scratch, using knowledge distillation (first and fifth rows). This approach, which is well-known and widely-employed, does not perform well in the few-shot setting, since there is not enough data to learn the underlying distribution. Next, we perform experiments using an ablation version of our method, which copies the weights from the teacher (second and sixth rows). This approach, which we refer to as Weight Copying + Knowledge Distillation (WeCo+KD), updates all the weights of the student. Even though this is a very strong baseline, reaching an accuracy of $53.7\%$ in the linear probing setting on ImageNet when $10\%$ of the initial training data is used during knowledge distillation, it is still $2.3\%$ behind our approach. Nevertheless, weight copying brings significant performance boosts over the standard distillation, confirming its practical utility.

The third ablated model uses weight copying and adds LoRA to the projection matrices corresponding to the query and value tokens $\boldsymbol{Q}$ and $\boldsymbol{V}$, following Luo \etal~\cite{luo2023lcm}. This approach performs poorly, reaching only $33.0\%$ in terms of accuracy on the ImageNet downstream task, when $10\%$ of the initial training data is employed. We conjecture that LoRA adds an insufficient number of learnable weights to properly learn the skipped layers of the teacher. Our full framework, WeCoLoRA, replaces standard LoRA with our enhanced version of LoRA, which is based on adding LoRA layers to each component of the transformer blocks. Our enhanced LoRA (fourth and eight rows in Table~\ref{tab:comp_to_kd}) outperforms both LoRA and WeCo+KD on all data sets. In summary, the ablation study demonstrates the utility of our novel components.


\noindent
\textbf{Comparison with state-of-the-art methods.} 
In Table~\ref{tab:comp_to_sota_kd}, we compare WeCoLoRA and WeCo+KD with four state-of-the-art techniques: DeiT \cite{Touvron-ICML-2021}, MiniViT \cite{Zhang-CVPR-2022}, DMAE \cite{bai2023masked} and DoRA \cite{Liu-ICML-2024}. We also report the performance obtained by the self-supervised~\cite{He-CVPR-2022} and supervised \cite{Dosovitskiy-ICLR-2021} teachers (both based on ViT-B) on each downstream task, as an upper bound for the few-shot methods. Note that MiniViT and DoRA are not directly applicable in conjunction with self-supervised teachers.

Interestingly, the ablated version of WeCoLoRA, namely WeCo+KD, generally outperforms all state-of-the-art methods~\cite{bai2023masked, Liu-ICML-2024, Touvron-ICML-2021,Zhang-CVPR-2022}. This indicates that current methods are not able to handle the limited amount of data that is typical to the few-shot setting. WeCoLoRA surpasses all four state-of-the-art KD methods~\cite{bai2023masked, Liu-ICML-2024, Touvron-ICML-2021,Zhang-CVPR-2022} by significant margins, \eg the differences are between $5\%$ and $45\%$ on ImageNet-1K. Moreover, on the iNauralist data set, our method obtains an accuracy that is only $3\%$ below the supervised teacher, while using half the number of layers and only using $10\%$ of the training data during distillation. Since WeCo+KD, the ablated version of our method, generally surpasses the state-of-the-art methods, we choose it as a strong baseline for the subsequent experiments.

\begin{table*}[t]
    \centering
    \small{
    \begin{tabular}{|c|c|c|c|c|c|c|c|}
        \hline
        \multirow{2}{*}{Teacher} & \multirow{1}{*}{Compression} & \multirow{1}{*}{Distillation} & \multirow{2}{*}{\rotatebox[origin=c]{0}{ImageNet}} & \multirow{2}{*}{\rotatebox[origin=c]{0}{ChestX-ray14}} & \multirow{2}{*}{\rotatebox[origin=c]{0}{iNaturalist}} & \multirow{2}{*}{\rotatebox[origin=c]{0}{RESISC45}} & \multirow{2}{*}{\rotatebox[origin=c]{0}{CIFAR-100}} \\
                & \multirow{1}{*}{factor $r$} & \multirow{1}{*}{method}            &           &     &      &        &    \\
        \hline
        \hline
        \multirow{3}{*}{ViT-B~\cite{Dosovitskiy-ICLR-2021}} 
        & \multirow{2}{*}{2} 
        & WeCo+KD & $35.5$ & $67.6$ & $25.4$ & $58.0$ & $34.8$  \\
        \multirow{3}{*}{(supervised)} & & WeCoLoRA & $\mathbf{60.5}$ & $\mathbf{69.5}$ & $\mathbf{43.4}$ & $\mathbf{67.5}$ & $\mathbf{60.6}$  \\ \cline{2-8}
        & \multirow{2}{*}{3}  & WeCo+KD & $7.6$ & $57.2$ &  $6.0$  & $33.3$ & $11.0$  \\ 
       &  &  WeCoLoRA &  $\mathbf{35.8}$ & $\mathbf{67.6}$ & $\mathbf{26.6}$   &   $\mathbf{58.8}$ & $\mathbf{36.5}$  \\
        \hline
        
        \multirow{3}{*}{ViT-B~\cite{He-CVPR-2022}}  
         & \multirow{2}{*}{2} & WeCo+KD 
        & $30.2$ & $65.7$ & $19.2$ & $53.8$ & $32.1$  \\ 
        
        \multirow{3}{*}{(SSL)} & & WeCoLoRA & $\mathbf{36.8}$ & $\mathbf{66.3}$ & $\mathbf{22.7}$ & $\mathbf{56.8}$ & $\mathbf{33.4}$ \\ \cline{2-8}

        & \multirow{2}{*}{3}  & WeCo+KD & $32.4$  & $\mathbf{66.7}$  &  $\mathbf{21.1}$ & $42.2$  & $34.0$  \\         
        &  & WeCoLoRA & $\mathbf{32.9}$  & $66.6$  & $\mathbf{21.1}$ & $\mathbf{55.6}$ & $\mathbf{34.4}$ \\ 
        \hline
    \end{tabular}
    }
        \vspace{-0.1cm}
        \caption{Results of WeCoLoRA and WeCo+KD in terms of accuracy (in percentages) on ImageNet-1K \cite{Deng-CVPR-2009}, iNaturalist \cite{VanHorn-CVPR-2018}, NWPU-RESISC45 \cite{Cheng-PIEEE-2017} and CIFAR-100 \cite{Krizhevsky-TR-2009}, and in terms of mean Area Under the Curve (AUC, in percentages) on ChestX-ray14 \cite{Wang-CVPR-2017}. Results are reported for two teachers: supervised ViT-B~\cite{Dosovitskiy-ICLR-2021} and self-supervised (SSL) ViT-B~\cite{He-CVPR-2022}.  During the distillation procedure, only $\mathbf{1\%}$ of the ImageNet-1K training set~\cite{Deng-CVPR-2009} is used. The best score on each data set for each teacher and each compression rate is highlighted in bold.}
    \label{tab:results_1}
        \vspace{-0.1cm}
\end{table*} 

\begin{table*}[!h]
    \centering
     \small{
    \begin{tabular}{|c|c|c|c|c|c|c|c|}
           \hline
        \multirow{2}{*}{Teacher} & \multirow{1}{*}{Compression} & \multirow{1}{*}{Distillation} & \multirow{2}{*}{\rotatebox[origin=c]{0}{ImageNet}} & \multirow{2}{*}{\rotatebox[origin=c]{0}{ChestX-ray14}} & \multirow{2}{*}{\rotatebox[origin=c]{0}{iNaturalist}} & \multirow{2}{*}{\rotatebox[origin=c]{0}{RESISC45}} & \multirow{2}{*}{\rotatebox[origin=c]{0}{CIFAR-100}} \\
                & \multirow{1}{*}{factor $r$} & \multirow{1}{*}{method}            &           &     &      &        &    \\
        \hline
        \hline
        \multirow{3}{*}{ViT-B~\cite{Dosovitskiy-ICLR-2021}} 
        & \multirow{2}{*}{2} 
        & WeCo+KD & $68.3$ & $69.8$ & $45.7$ & $\mathbf{74.0}$ & $64.4$   \\ 
        \multirow{3}{*}{(supervised)} & & WeCoLoRA & $\mathbf{69.2}$ & $\mathbf{70.0}$ & $\mathbf{49.5}$ & ${70.5}$ & $\mathbf{68.3}$   \\ 
        \cline{2-8}  
         & \multirow{2}{*}{3} & WeCo+KD & $58.1$ & $68.4$ & $38.3$ & $\mathbf{67.7}$ & $51.4$  \\       
         & & WeCoLoRA & $\mathbf{58.3}$ & $\mathbf{69.1}$ & $\mathbf{41.6}$ & $66.8$ & $\mathbf{54.1}$ \\ 
         \hline
        
        \multirow{3}{*}{ViT-B~\cite{He-CVPR-2022}}  
         & \multirow{2}{*}{2}
         & WeCo+KD & $53.7$ & $69.7$ & $31.2$ & $61.4$ & $48.2$   \\  
        \multirow{3}{*}{(SSL)} & & WeCoLoRA 
        & $\mathbf{56.0}$ & $\mathbf{69.8}$ & $\mathbf{34.5}$ & $\mathbf{62.3}$ & $\mathbf{49.8}$    \\ 
        \cline{2-8}

        & \multirow{2}{*}{3} & WeCo+KD & $40.6$ & $\mathbf{67.6}$ & $22.7$  & $53.5$  & $34.6$  \\
        
        & & WeCoLoRA & $\mathbf{41.0}$ & $67.3$ & $\mathbf{23.5}$ & $\mathbf{60.2}$ & $\mathbf{39.5}$   \\ \hline
    \end{tabular} 
    }
        \vspace{-0.3cm}
            \caption{Results of WeCoLoRA and WeCo+KD in terms of accuracy (in percentages) on ImageNet-1K \cite{Deng-CVPR-2009}, iNaturalist \cite{VanHorn-CVPR-2018}, NWPU-RESISC45 \cite{Cheng-PIEEE-2017} and CIFAR-100 \cite{Krizhevsky-TR-2009}, and in terms of mean Area Under the Curve (AUC, in percentages) on ChestX-ray14 \cite{Wang-CVPR-2017}. Results are reported for two teachers: supervised ViT-B~\cite{Dosovitskiy-ICLR-2021} and self-supervised (SSL) ViT-B~\cite{He-CVPR-2022}.  During the distillation procedure, only $\mathbf{10\%}$ of the ImageNet-1K training set~\cite{Deng-CVPR-2009} is used. The best score on each data set for each teacher and each compression rate is highlighted in bold.}\label{tab:results_10}
        \vspace{-0.3cm}
\end{table*} 

\noindent
\textbf{Results with multiple teachers and compression rates.}
We report additional results with WeCoLoRA and its strong ablated version, WeCo+KD, in Tables~\ref{tab:results_1} and \ref{tab:results_10}. We alternatively employ two teachers in these experiments, a supervised ViT-B~\cite{Dosovitskiy-ICLR-2021} and a self-supervised ViT-B~\cite{He-CVPR-2022}. We also consider two compression factors, namely $r=2$ and $r=3$, on five downstream tasks.

In Table~\ref{tab:results_1}, we report results when only $1\%$ of the ImageNet-1K training set~\cite{Deng-CVPR-2009} is employed during the knowledge distillation procedure. In this scenario, WeCoLoRA outperforms the baseline in most cases. Remarkably, for $r=2$, WeCoLoRA achieves an accuracy of $60.5\%$ on ImageNet using only $1\%$ of the training data, surpassing the accuracy of WeCo+KD by $25\%$.

We present results for $10\%$ of the distillation data in Table~\ref{tab:results_10}. Once again, WeCoLoRA outperforms the baseline in most settings. However, the gains in favor of our method when using $10\%$ of the data (Table~\ref{tab:results_10}) are generally lower than distilling on $1\%$ of the data (Table~\ref{tab:results_1}). We consider that $10\%$ of ImageNet-1K, \ie about 100K training images, is at the upper end of the scenarios that can still be regarded as \emph{few-shot}. The higher number of trainable parameters in WeCo+KD, combined with the large training set, allow it to recover the gap with respect to WeCoLoRA. To further demonstrate the superiority of WeCoLoRA in other few-shot scenarios, we present additional results when using $0.25\%$, $2\%$ and $5\%$ of the data during distillation, in the supplementary.

\begin{figure}[t]
  \centering
  \includegraphics[width=0.8\linewidth]{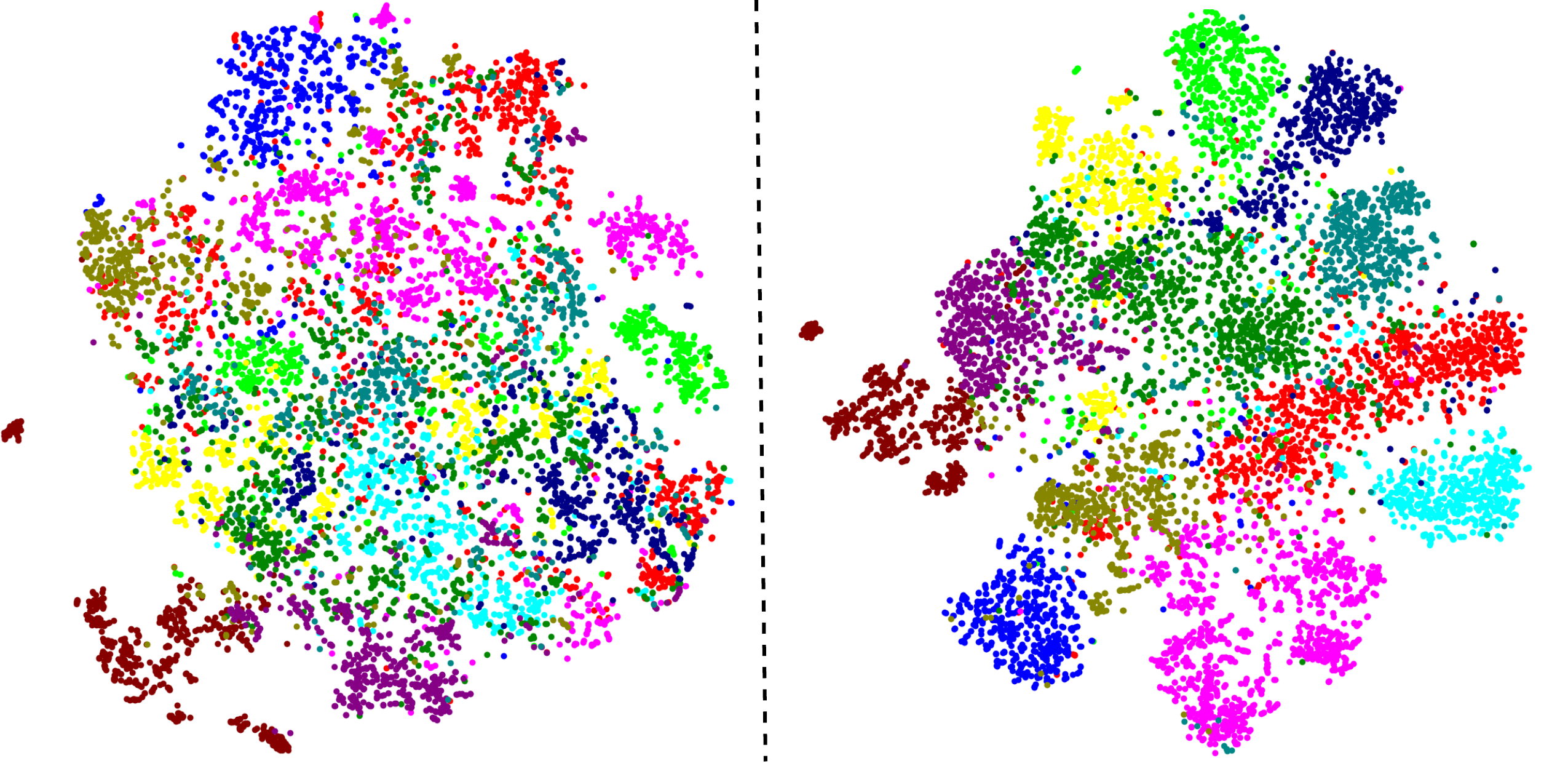}
      \vspace{-0.25cm}
  \caption{Visualizations of the latent spaces learned by WeCo+KD (on the left-hand side) and our method (on the right-hand side). Both visualizations are obtained with t-SNE. The embeddings correspond to images from the RESISC45~\cite{Cheng-PIEEE-2017} test set, before linear probing. The features are extracted from student models that are distilled from the supervised teacher ViT-B~\cite{He-CVPR-2022}, using a compression factor of $r=2$. During the distillation procedure, only $1\%$ of ImageNet data set~\cite{Deng-CVPR-2009} is used. The colors correspond to the class labels from RESISC45. Best viewed in color.}  
  \label{fig_tsne}
      \vspace{-0.4cm}
\end{figure}

\noindent
\textbf{Latent space visualizations.} 
In Figure~\ref{fig_tsne}, we illustrate t-SNE visualizations of the embeddings learned with the proposed method (WeCoLoRA) versus its ablated version denoted as WeCo+KD. The embeddings are obtained on the RESISC45~\cite{Cheng-PIEEE-2017} test set. We illustrate the embeddings of the student obtained through distilling the supervised ViT-B~\cite{Dosovitskiy-ICLR-2021} teacher, with a compression ratio of $r=2$. We notice that our WeCoLoRA is able to learn a discriminative feature distribution during distillation, disentangling the out-of-domain (downstream) data samples from RESISC45 into clearly delineated clusters, even though there was no supervision involved, other than the teacher features. However, WeCo+KD, which is based on complete weight adaptation during distillation, creates clusters which are spread across the feature space. In summary, the t-SNE visualizations emphasize that WeCoLoRA is an effective few-shot knowledge distillation approach, explaining the superior results on downstream linear probing tasks through the robustness of the learned latent space.

\begin{figure}[t]
  \centering
  \includegraphics[width=0.65\linewidth]{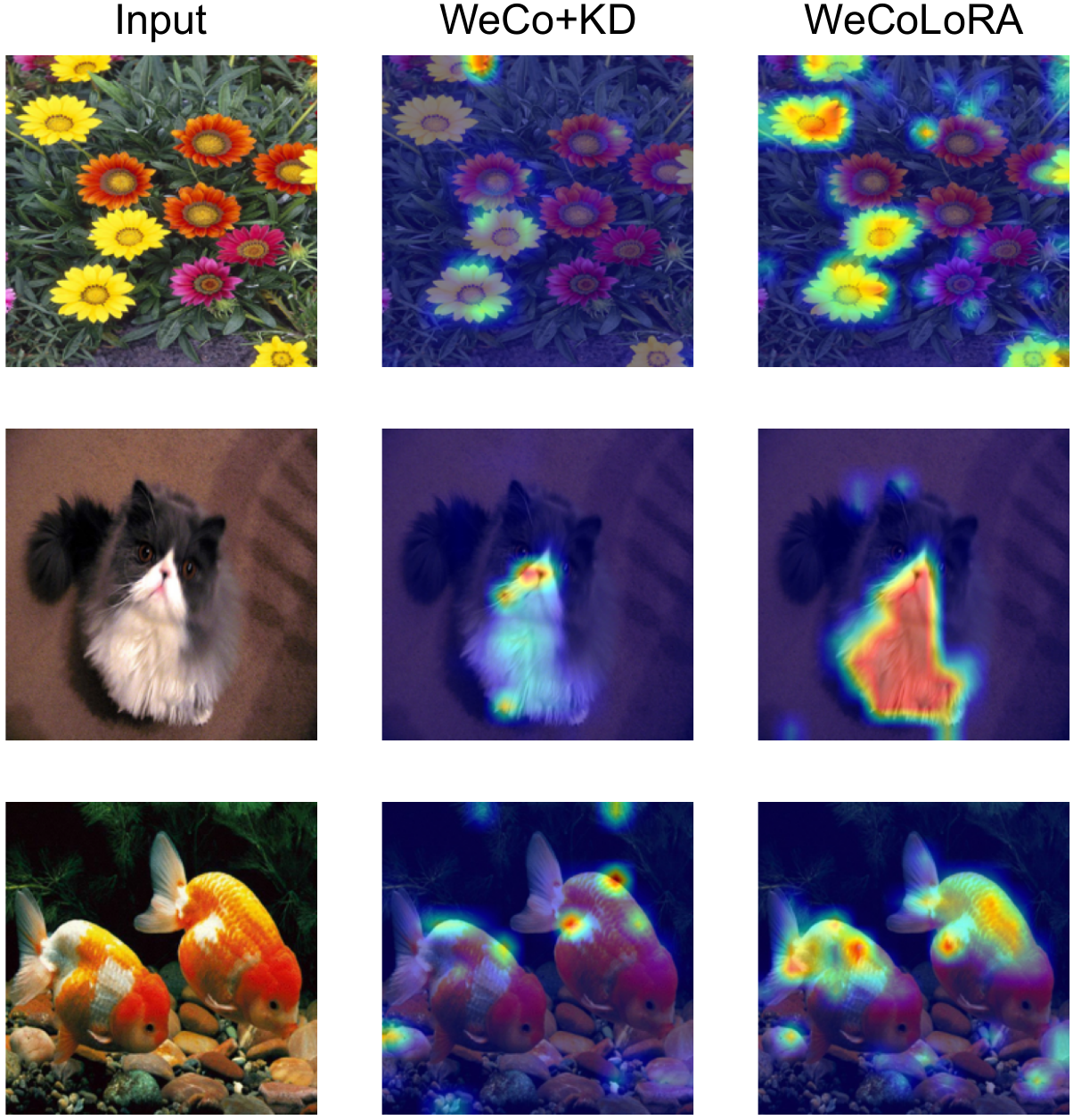}
    \vspace{-0.2cm}
  \caption{Attention visualizations obtained with Attention Rollout \cite{abnar2020quantifying} for WeCo+KD and WeCoLoRA, on three test images from ImageNet-1K. The compared students are distilled from the supervised ViT-B teacher \cite{Dosovitskiy-ICLR-2021}, with a compression factor of $r=2$, on $1\%$ of the original training data. The first column displays the original images, the second column displays the attention of the WeCo+KD-based student, and the third column shows the attention of the WeCoLoRA-based student. Best viewed in color.
}  
  \label{fig_attention}
    \vspace{-0.4cm}
\end{figure}

\noindent
\textbf{Attention visualization.} To visualize the attention, we employ Attention Rollout \cite{abnar2020quantifying}, a technique that enables the visualization of the attention flow throughout ViT. We fuse the attention heads by taking the maximum response (top $10\%$) and discard pixels with lower values. In Figure \ref{fig_attention}, we present three test images that are randomly taken from ImageNet \cite{Deng-CVPR-2009}, which correspond to ``Daisy'', ``Persian Cat'' and  
``Goldfish'' classes, respectively. The attention is extracted from student models trained with WeCo+KD and WeCoLoRA. We observe that WeCoLoRA makes the model focus more on discriminative features such as petals, fur, and fish scales. This helps the model based on WeCoLoRA in taking informed decisions.

\begin{figure}[t]
  \centering
  \includegraphics[width=0.55\linewidth]{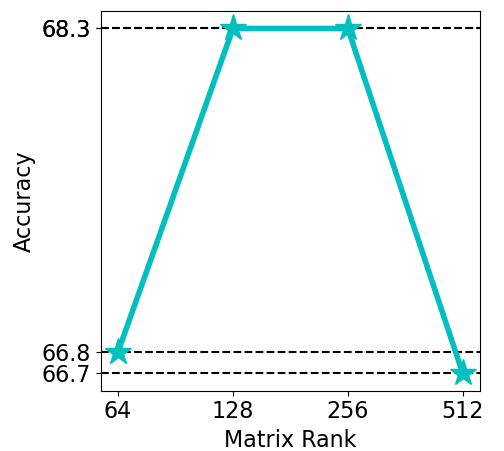}
  \vspace{-0.15cm}
  \caption{Accuracy rates of WeCoLoRA on the CIFAR-100 data set~\cite{Krizhevsky-TR-2009} when varying the matrix rank. The teacher is the supervised ViT-B~\cite{Dosovitskiy-ICLR-2021}, the compression factor is $r=2$, and the distillation process uses $10\%$ of the ImageNet-1K training set~\cite{Deng-CVPR-2009}.}  
  \vspace{-0.4cm}
  \label{fig_rank}
\end{figure}

\noindent
\textbf{Varying the matrix rank.} In Figure~\ref{fig_rank}, we present results obtained by WeCoLoRA on the CIFAR-100 data set, when changing the matrix rank $k \in \{64, 128, 256, 512 \}$. The teacher is the supervised ViT-B~\cite{Dosovitskiy-ICLR-2021}, the compression factor is $r=2$, and the distillation process is based on $10\%$ of the original training data. We observe that WeCoLoRA obtains stable performance when $k=128$ and $k=256$. If $k$ is too small ($k=64$), the performance drops, showing that the model needs more parameters to recover the information of the skipped teacher layers. However, the performance also drops when $k$ is too large ($k=512$), but this happens because the model starts to overfit the small data set. Hence, we observe that $k$ is a hyperparameter that can control the bias-variance trade-off of WeCoLoRA. In our experiments, we found that $k=128$ and $k=256$ perform generally well in the few-shot distillation scenarios.

\noindent
\textbf{Computational evaluation.}  
We first compare the running time of the teacher ViT-B and the student based on a compression ratio $r=2$. As expected, the student is twice as fast as its teacher, evaluating $512$ samples in $3$ \texttt{ms}, while the teacher model requires $6$ \texttt{ms} for the same mini-batch size. Next, we compare WeCoLoRA and WeCo+KD in terms of the number of trainable parameters. WeCo+KD updates $44$M parameters, while WeCOLoRA updates only $10$M parameters (when $k=128$). Because the new weights added by WeCoLoRA are not integrated into the network during training, WeCoLoRA uses $10.6$ GFLOPs, while WeCo+KD utilizes $8.5$ GFLOPs. During inference, the GFLOPs are the same for both WeCoLoRA and WeCo+KD. The reported running times are measured on a GeForce RTX 3090 GPU with 24G of VRAM.

\vspace{-0.1cm}
\section{Conclusion}
\label{sec:conclusions} 
\vspace{-0.1cm}

In this paper, we proposed a novel few-shot unsupervised feature distillation method that can be used to train vision transformers on a few unlabeled images. Our approach, termed WeCoLoRA, combines weight copying and low-rank adaptation in an efficient and effective training pipeline. We conducted experiments in multiple few-shot scenarios, using both supervised and self-supervised teachers, and considering various model compression factors. The results show that our method outperforms state-of-the-art competitors \cite{bai2023masked,Touvron-ICML-2021}, as well as ablated versions of our approach. Moreover, we present feature visualizations that clearly indicate that WeCoLoRA produces more robust embeddings, which are able to better disentangle the classes, even on downstream data sets.


\section*{Acknowledgments}
This work was supported by a grant of the Ministry of Research, Innovation and Digitization, CCCDI - UEFISCDI, project number PN-IV-P6-6.3-SOL-2024-2-0227, within PNCDI IV. The research leading to these results has also received funding from the NO Grants 2014-2021, under project ELO-Hyp contract no. 24/2020.



{\small
\bibliographystyle{ieee_fullname}
\bibliography{main}
} 

\clearpage
{\centering
        \Large
        \vspace{0.5em}\textbf{Supplementary Material} \\
        \vspace{1.0em}
}

\section{Additional Results}

\noindent
\textbf{Few-shot distillation from 1\% to 10\%.}
In Tables~\ref{tab:results_2} and \ref{tab:results_5}, we report results on five downstream tasks, when the students use only $2\%$ and $5\%$ of the ImageNet data during distillation, respectively.  As noted in the main manuscript, we perform linear probing to demonstrate that our method transfers strong features.
We notice that our method, WeCoLoRA, attains higher performance than the WeCo+KD method, especially when the distillation data is scarce. We observe a substantial improvement (of at least $2\%$) on the ImageNet downstream task, regardless of the reduction ratio or the distillation training size, when the teacher is the supervised ViT-B~\cite{Dosovitskiy-ICLR-2021} model. We observe the same trend on the other data sets employed in the evaluation. We further note that the features learned by our distillation method also transfer to out-of-distribution data sets, such as ChestX-ray14~\cite{Wang-CVPR-2017}. We consider ChestX-ray14 as out-of-distribution because it contains medical images, while the pre-training data set, ImageNet, contains natural images. 

We conclude that the proposed distillation method, WeCoLoRA, is robust and obtains improved performance on multiple downstream tasks, especially when the pre-training data set is small. We also emphasize that our method does not require labeled data, and is able to compress both supervised and self-supervised models.

\begin{table*}[!t]
    \centering
    \begin{tabular}{|c|c|c|c|c|c|c|c|}
           \hline
        \multirow{5}{*}{Teacher} &  \multirow{4}{*}{Compression} & \multirow{4}{*}{Distillation} & \multirow{5}{*}{\rotatebox[origin=c]{45}{ImageNet}} & \multirow{5}{*}{\rotatebox[origin=c]{45}{ChestX-ray14}} & \multirow{5}{*}{\rotatebox[origin=c]{45}{iNaturalist}} & \multirow{5}{*}{\rotatebox[origin=c]{45}{RESISC45}} & \multirow{5}{*}{\rotatebox[origin=c]{45}{CIFAR-100}} \\
                & \multirow{4}{*}{factor $r$} & \multirow{4}{*}{method}                       &     &      &        &  &  \\
                        &       &           &     &      &        &    &\\
                        &       &           &     &      &        &    &\\
                       &       &           &     &      &        &   & \\
        \hline
        \hline
         
          \multirow{3}{*}{ViT-B~\cite{Dosovitskiy-ICLR-2021}}  & \multirow{2}{*}{2}  & WeCo+KD & $46.9$ & $68.3$ & $35.1$ & $61.6$ &   $42.0$  \\ 
           & & WeCoLoRA &  $\mathbf{63.5}$ & $\mathbf{70.0}$ & $\mathbf{46.5}$ & $\mathbf{68.5}$ & $\mathbf{62.9}$  \\ 
         \cline{2-8}  
          (supervised) &
          \multirow{2}{*}{3} & WeCo+KD &  $37.0$ & $67.8$  & $28.2$ & $59.8$ & $37.9$ \\       
          & & WeCoLoRA &   $\mathbf{39.6}$ & $\mathbf{68.1}$ & $\mathbf{29.5}$ & $\mathbf{61.5}$ & $\mathbf{38.6}$ \\ 
         \hline

        \multirow{3}{*}{ViT-B~\cite{He-CVPR-2022}}  & \multirow{2}{*}{2}  & WeCo+KD & $46.7$ & $68.6$ & $28.3$ & $\mathbf{62.9}$ &   $41.5$  \\ 
           & & WeCoLoRA &  $\mathbf{48.2}$ & $\mathbf{68.9}$ & $\mathbf{28.5}$ & $58.8$ & $\mathbf{44.8}$  \\ 
         \cline{2-8}  
          (SSL) &
          \multirow{2}{*}{3} & WeCo+KD &  $33.6$ & $66.5$  & $20.0$ & $53.7$ & $35.7$ \\       
          & & WeCoLoRA &   $\mathbf{35.3}$ & $\mathbf{67.0}$ & $\mathbf{22.2}$ & $\mathbf{56.0}$ & $\mathbf{36.8}$ \\ 
         \hline
        
    \end{tabular} 
        \vspace{-0.2cm}
        \caption{Results of WeCoLoRA and WeCo+KD in terms of accuracy (in percentages) on ImageNet-1K \cite{Deng-CVPR-2009}, iNaturalist \cite{VanHorn-CVPR-2018}, NWPU-RESISC45 \cite{Cheng-PIEEE-2017} and CIFAR-100 \cite{Krizhevsky-TR-2009}, and in terms of mean AUC (in percentages) on ChestX-ray14 \cite{Wang-CVPR-2017}. Results are reported for the supervised ViT-B~\cite{Dosovitskiy-ICLR-2021} teacher and the self-supervised (SSL) ViT-B~\cite{He-CVPR-2022} teacher.  During the distillation procedure, only $\mathbf{2\%}$ of the ImageNet-1K training set~\cite{Deng-CVPR-2009} is used.}\label{tab:results_2}

\end{table*} 

\begin{table*}[!t]
    \centering

    \begin{tabular}{|c|c|c|c|c|c|c|c|}
           \hline
         \multirow{5}{*}{Teacher} & \multirow{4}{*}{Compression} & \multirow{4}{*}{Distillation} & \multirow{5}{*}{\rotatebox[origin=c]{45}{ImageNet}} & \multirow{5}{*}{\rotatebox[origin=c]{45}{ChestX-ray14}} & \multirow{5}{*}{\rotatebox[origin=c]{45}{iNaturalist}} & \multirow{5}{*}{\rotatebox[origin=c]{45}{RESISC45}} & \multirow{5}{*}{\rotatebox[origin=c]{45}{CIFAR-100}}  \\
                & \multirow{4}{*}{factor $r$} & \multirow{4}{*}{method}            &    &      &        &  &  \\
                        &       &           &     &      &        &    &\\
                        &       &           &     &      &        &   & \\
                       &       &           &     &      &        &   & \\
        \hline
        \hline
         
         \multirow{3}{*}{ViT-B~\cite{Dosovitskiy-ICLR-2021}}  &  \multirow{2}{*}{2}  & WeCo+KD & $65.3$ & $69.4$ & $45.4$ & $\mathbf{73.5}$ & $60.1$\\ 
          &  & WeCoLoRA &  $\mathbf{67.3}$ &  $\mathbf{70.0}$ & $\mathbf{49.0}$ & $69.9$ & $\mathbf{66.6}$\\ 
         \cline{2-8}  
             (supervised) & \multirow{2}{*}{3} & WeCo+KD & $52.2$ & $68.4$ &  $37.0$ &   $\mathbf{66.7}$ & $46.9$\\       
          & & WeCoLoRA &   $\mathbf{55.3}$ & $\mathbf{69.9}$ & $\mathbf{40.2}$ & $66.1$ & $\mathbf{51.7}$\\ 
         \hline

          \multirow{3}{*}{ViT-B~\cite{He-CVPR-2022}}  &  \multirow{2}{*}{2}  & WeCo+KD & $51.0$ & $69.3$ & $29.9$ & $\mathbf{61.7}$ & $47.6$\\ 
          &  & WeCoLoRA &  $\mathbf{54.0}$ &  $\mathbf{69.5}$ & $\mathbf{32.9}$ & $61.5$ & $\mathbf{47.7}$\\ 
         \cline{2-8}  
          (SSL) & \multirow{2}{*}{3} &  WeCo+KD & $36.1$ & $66.6$ &  $18.7$ &   $51.3$ & $30.5$\\       
          & & WeCoLoRA  &   $\mathbf{37.4}$ & $\mathbf{66.5}$ & $\mathbf{22.0}$ & $\mathbf{58.6}$ & $\mathbf{38.6}$\\ 
         \hline
        
    \end{tabular} 
        \vspace{-0.2cm}
        \caption{Results of WeCoLoRA and WeCo+KD in terms of accuracy (in percentages) on ImageNet-1K \cite{Deng-CVPR-2009}, iNaturalist \cite{VanHorn-CVPR-2018}, NWPU-RESISC45 \cite{Cheng-PIEEE-2017} and CIFAR-100 \cite{Krizhevsky-TR-2009}, and in terms of mean AUC (in percentages) on ChestX-ray14 \cite{Wang-CVPR-2017}. Results are reported for the supervised ViT-B~\cite{Dosovitskiy-ICLR-2021} teacher and the self-supervised (SSL) ViT-B~\cite{He-CVPR-2022} teacher.  During the distillation procedure, only $\mathbf{5\%}$ of the ImageNet-1K training set~\cite{Deng-CVPR-2009} is used.}\label{tab:results_5}
        \vspace{-0.1cm}
\end{table*} 

\begin{figure*}[!h]
  \centering
  \includegraphics[width=0.7\textwidth]{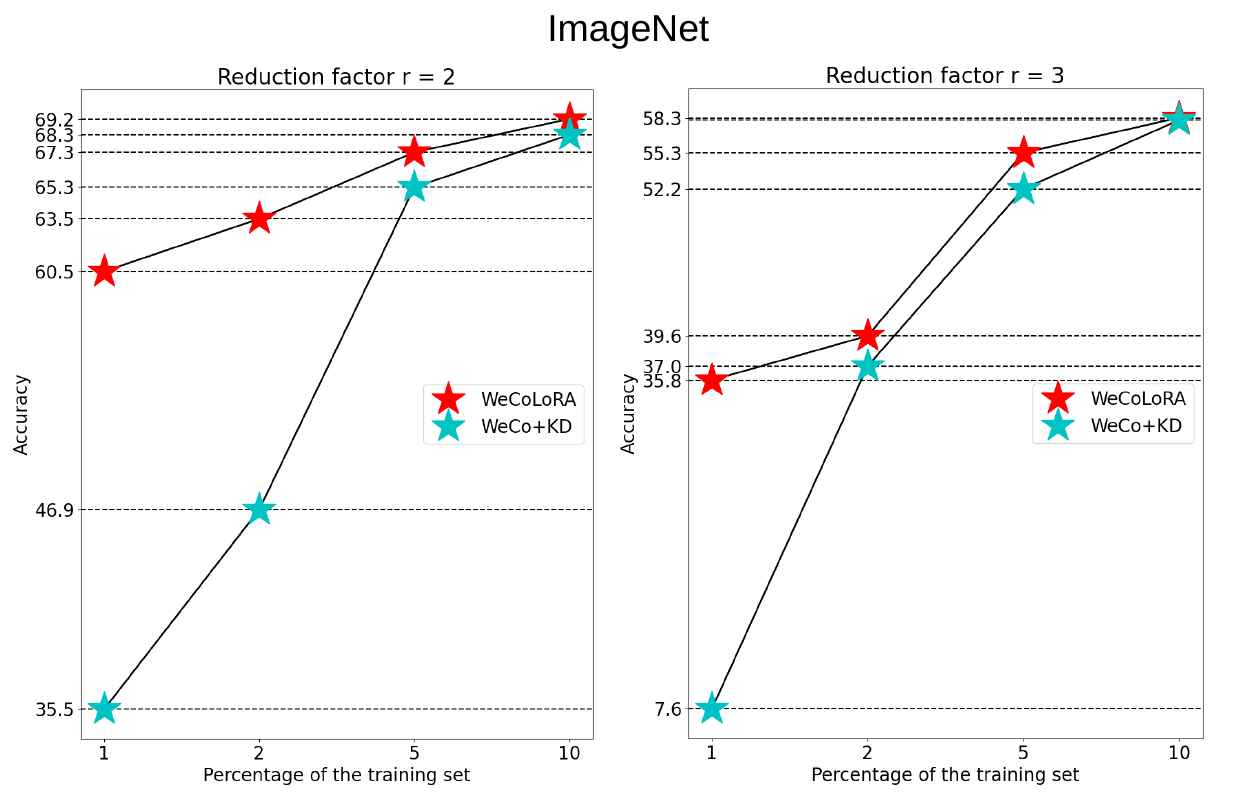}
      \vspace{-0.2cm}
  \caption{Accuracy rates obtained by WeCoLoRA and WeCo+KD on the ImageNet-1K~\cite{Deng-CVPR-2009} downstream task.
Results are reported for the supervised  ViT-B~\cite{Dosovitskiy-ICLR-2021} teacher. The horizontal axis corresponds to the percentage of the original training set~\cite{Deng-CVPR-2009} used during knowledge distillation.  Best viewed in color.}  
  \label{fig_imagenet}
      \vspace{-0.1cm}
\end{figure*}
\begin{figure*}[!h]
  \centering
  \includegraphics[width=0.7\textwidth]{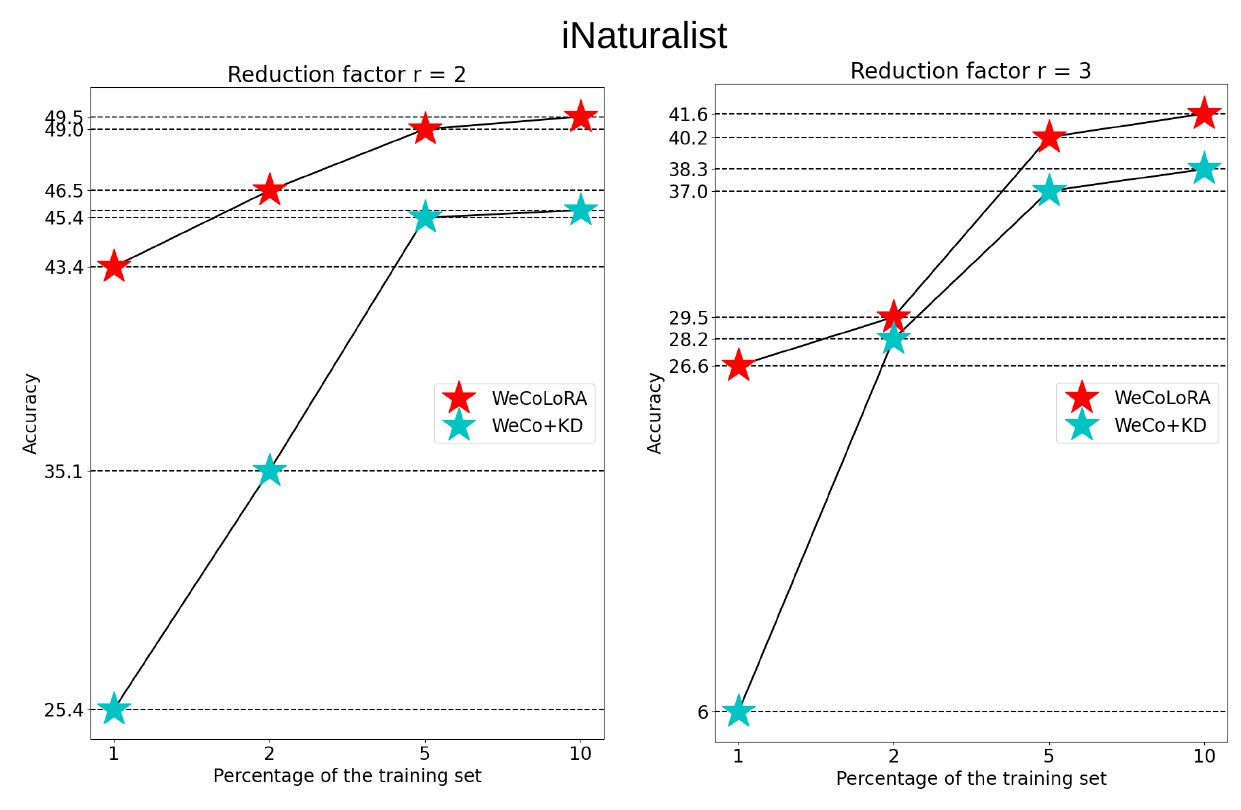}
      \vspace{-0.2cm}
  \caption{ Accuracy rates obtained by WeCoLoRA and WeCo+KD on the iNaturalist~\cite{VanHorn-CVPR-2018} downstream task.
Results are reported for the supervised  ViT-B~\cite{Dosovitskiy-ICLR-2021} teacher. The horizontal axis corresponds to the percentage of the original training set~\cite{Deng-CVPR-2009} used during knowledge distillation. Best viewed in color.}  
  \label{fig_inat}
      \vspace{-0.3cm}
\end{figure*}
\begin{figure*}[!h]
  \centering
  \includegraphics[width=0.7\textwidth]{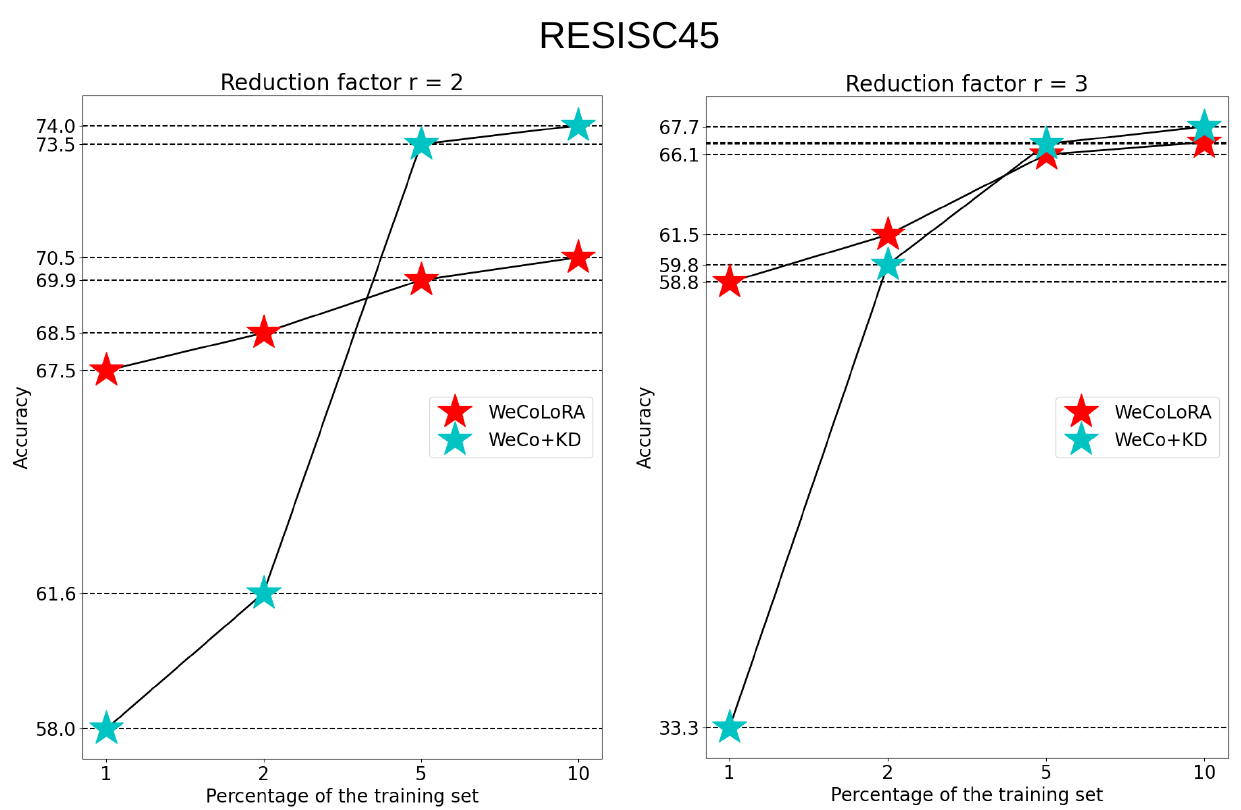}
      \vspace{-0.1cm}
  \caption{Accuracy rates obtained by WeCoLoRA and WeCo+KD on the NWPU-RESISC45 \cite{Cheng-PIEEE-2017} downstream task.
Results are reported for the supervised  ViT-B~\cite{Dosovitskiy-ICLR-2021} teacher. The horizontal axis corresponds to the percentage of the original training set~\cite{Deng-CVPR-2009} used during knowledge distillation. Best viewed in color.}  
  \label{fig_nw}
      \vspace{-0.1cm}
\end{figure*}
\begin{figure*}[!h]
  \centering
  \includegraphics[width=0.7\textwidth]{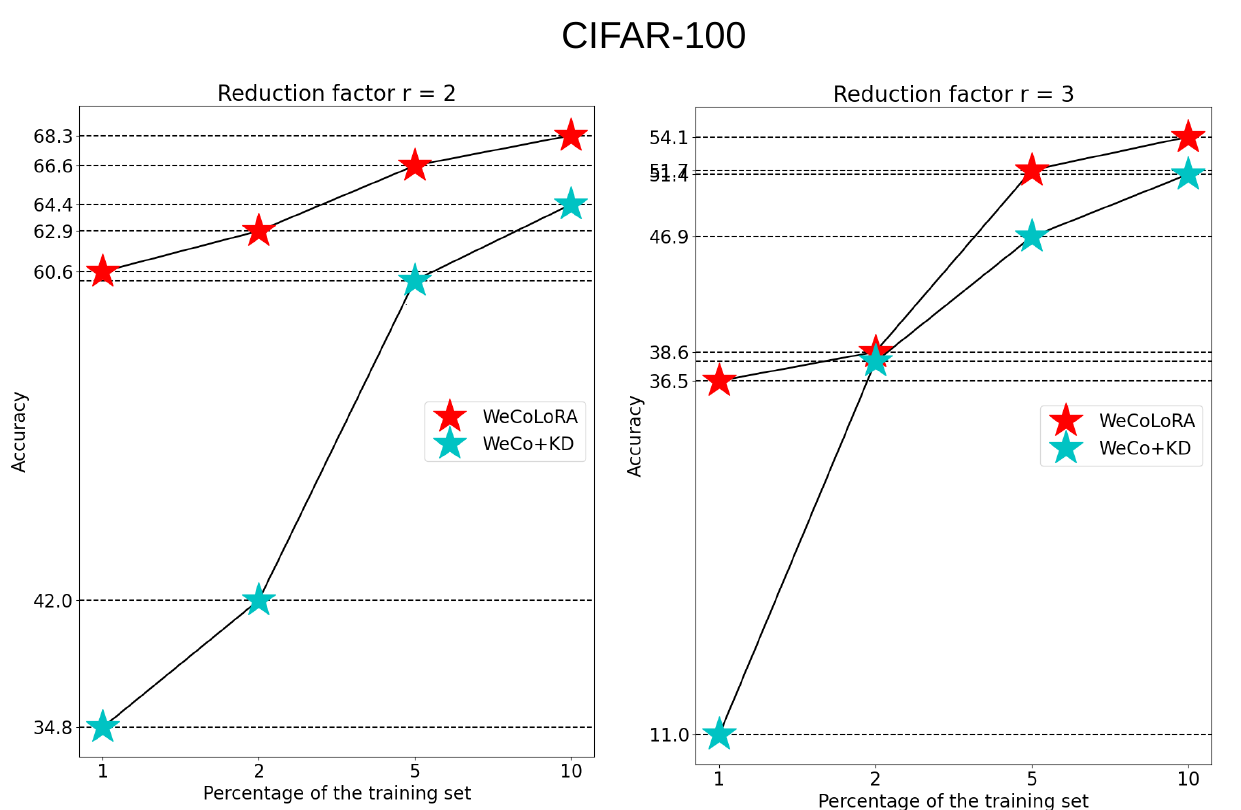}
      \vspace{-0.1cm}
  \caption{Accuracy rates obtained by WeCoLoRA and WeCo+KD on the CIFAR-100 \cite{Krizhevsky-TR-2009} downstream task.
Results are reported for the supervised  ViT-B~\cite{Dosovitskiy-ICLR-2021} teacher. The horizontal axis corresponds to the percentage of the original training set~\cite{Deng-CVPR-2009} used during knowledge distillation. Best viewed in color.}  
  \label{fig_cifar}
      \vspace{-0.3cm}
\end{figure*}
\begin{figure*}[!h]
  \centering
  \includegraphics[width=0.7\textwidth]{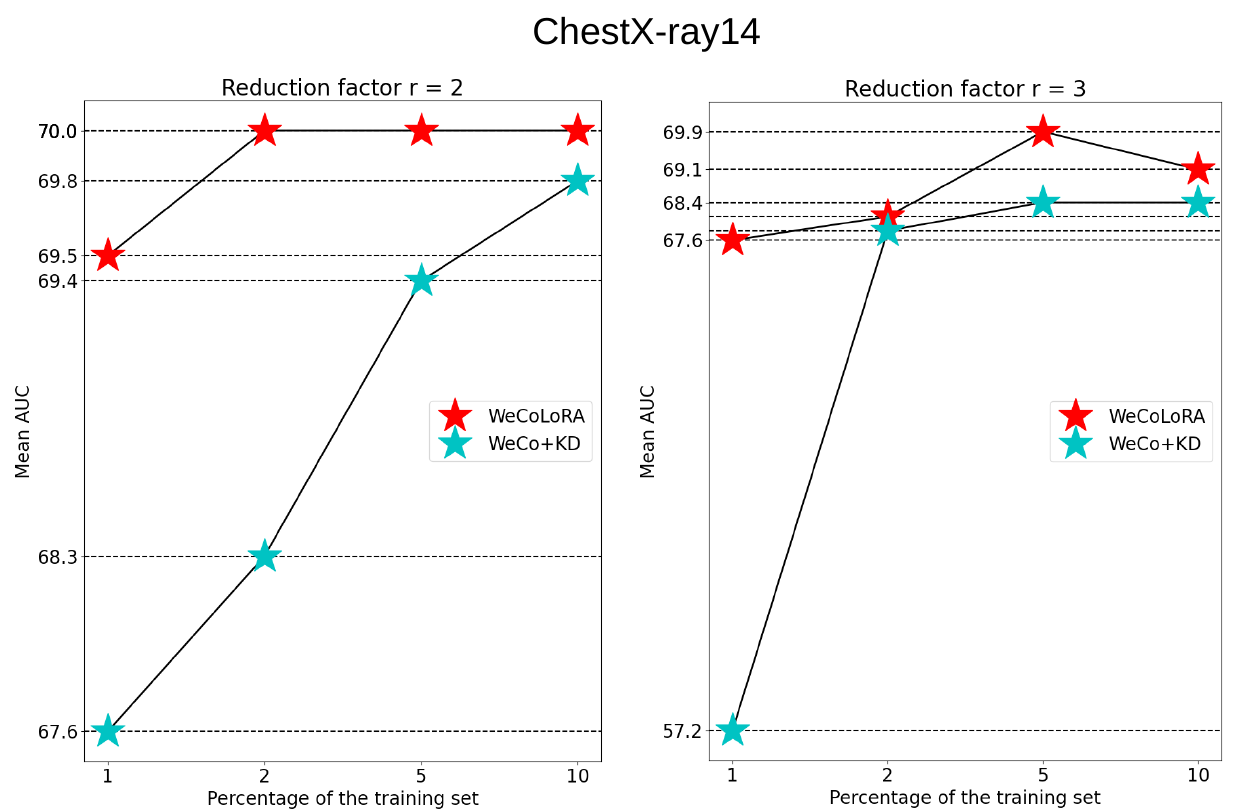}
    \vspace{-0.2cm}
  \caption{Mean AUC scores (in percentages) obtained by WeCoLoRA and WeCo+KD on the ChestX-ray14 \cite{Wang-CVPR-2017} downstream task.
Results are reported for the supervised  ViT-B~\cite{Dosovitskiy-ICLR-2021} teacher. The horizontal axis corresponds to the percentage of the original training set~\cite{Deng-CVPR-2009} used during knowledge distillation. Best viewed in color.}  
  \label{fig_xray}
      \vspace{-0.1cm}
\end{figure*}

To better assess the performance trends on various downstream tasks when the number of samples increases from $1\%$ to $10\%$, we further illustrate the performance levels obtained by WeCoLoRA vs.~WeCo+KD on ImageNet-1K \cite{Deng-CVPR-2009}, iNaturalist \cite{VanHorn-CVPR-2018}, NWPU-RESISC45 \cite{Cheng-PIEEE-2017}, CIFAR-100 \cite{Krizhevsky-TR-2009} and ChestX-ray14 \cite{Wang-CVPR-2017} in Figures~\ref{fig_imagenet}, \ref{fig_inat}, \ref{fig_nw}, \ref{fig_cifar}, and \ref{fig_xray}, respectively. We observe that WeCoLoRA obtains significantly higher performance than WeCo+KD when there is less data involved in the knowledge distillation process ($1\%$ and $2\%$ of the original training set~\cite{Deng-CVPR-2009}). Moreover, in most of the cases, WeCoLoRA also outperforms WeCo+KD when $10\%$ of the original training set in used during knowledge distillation.  We also conducted experiments on the full scale ImageNet and observed marginal differences between WeCo+KD and WeCoLoRA. We therefore conclude that WeCoLoRA is particularly useful in the few-shot KD setting.

\begin{table*}[t!]
    \centering
    \begin{tabular}{|l|c|c|c|}
    \hline  
         Method & ImageNet & iNaturalist & CIFAR-100 \\
        \hline   
        \hline
        WeCo+KD (ablated) & $14.0$ &  $9.9$ & $17.1$ \\
        WeCoLoRA (ours) & $21.6$ &  $14.6$ & $\textbf{25.7}$  \\
        WeCoLoRA (ours) + k-means++ init & $\textbf{21.7}$ &  $\textbf{14.9}$ & $25.5$  \\
        \hline  
    \end{tabular}
         \vspace{-0.2cm}
     \caption{Accuracy rates on ImageNet-1K, iNaturalist19 and CIFAR-100, when only 0.25\% ($\alpha\!=\!0.25$) of the training set is used during distillation. K-means++ init is used to select the  samples for few-shot KD (to avoid collapse). Compression factor: $r\!=\!4$. Teacher:  self-supervised ViT-B.}\label{tab:alpha_02}
        \vspace{-0.3cm}
\end{table*} 

\begin{table}[t!]
    \centering
    \setlength\tabcolsep{2.8pt}
    \begin{tabular}{|l|c|c|c|}
    \hline  
         Method & ImageNet & iNaturalist & CIFAR-100 \\
        \hline    
        \hline
        TopPruning \cite{Sajjad-CSL-2023} & $52.6$ &  $41.4$ & $56.1$  \\
        WeCoLoRA (ours) & $\textbf{69.2}$ & $\textbf{49.5}$ & $\textbf{68.3}$\\
        
        \hline  
    \end{tabular}
         \vspace{-0.2cm}
     \caption{Accuracy rates on ImageNet-1K, iNaturalist19 and CIFAR-100, comparing our WeCoLoRA with top-layer dropping (TopPruning). Compression factor: $r\!=\!2$. Teacher:  supervised ViT-B.}\label{tab:pruning}
        \vspace{-0.3cm}
\end{table} 

\begin{table}[t!]
    \centering
    \setlength\tabcolsep{2.5pt}
    \begin{tabular}{|l|c|c|c|c|}
    \hline  
         Method & Layer 1 & Layer 2 & Layer 3 & Layer 4 \\
        \hline   
        \hline
        WeCo (ablated) & $0.693$ &  $0.458$ & $0.387$ & $0.268$ \\
        WeCoLoRA (ours) & $0.753$ &  $0.686$ & $0.674$ & $0.796$  \\
        \hline  
    \end{tabular}
         \vspace{-0.2cm}
     \caption{Mean cosine similarities (averaged over tokens) between the student's layers and the skipped layers from the teacher. The values are calculated on features derived from 10\% of ImageNet. Compression factor: $r=3$. Teacher: self-supervised ViT-B.}
     \label{tab:cos}
        \vspace{-0.6cm}
\end{table}

\noindent
\textbf{Fewer samples, lighter student.}
We would like to mention that $1\%$ of ImageNet corresponds to 12 samples per class. However, models are often evaluated on even fewer shots. To this end, we perform extra experiments for $0.25\%$ of ImageNet, \ie~3 samples per class. In this setting, we also employ a more aggressive reduction ratio ($r=4$), which leads to a very light student model.

When the number of shots is very small, the model can collapse due to a low diversity of training samples. Although this is not the particular focus of our method, there is no obvious reason for WeCoLoRA not to be compatible with orthogonal methods that deal with the issue of collapse. To mitigate the issue of collapse, we combine WeCoLoRA with k-means++ init to select the training samples.

The results of WeCo+KD and WeCoLoRA for $0.25\%$ of ImageNet and $r=4$ are shown in Table \ref{tab:alpha_02}. WeCoLoRA obtains superior results on all three data sets (ImageNet, iNaturalist and CIFAR-100). When the training samples are chosen via k-means++, we obtain slightly improved results on two datasets (see the last row in Table \ref{tab:alpha_02}).

\noindent
\textbf{Layer pruning vs.~WeCoLoRA.}
A promising approach to create lighter models without much effort is layer pruning. For transformer models, it was recently found that TopPruning, a method that drops the top-layers, obtains surprisingly good results \cite{Sajjad-CSL-2023}. To this end, we compare WeCoLoRA with TopPruning, using the same reduction factor of $r=2$ for both methods. The results, which are reported in Table \ref{tab:pruning}, clearly indicate that WeCoLoRA outperforms TopPruning.

\noindent
\textbf{Comparing teacher and student features.} One question that arises when applying WeCoLoRA is if the student is indeed learning  features similar to the skipped teacher layers. To address this point, we compute the mean cosine similarities between the skipped layers and the corresponding student layers (from 1 to 4), before and after applying our enhanced LoRA. As shown in Table \ref{tab:cos}, the similarities increase after distillation, indicating that the student learns features similar to the skipped teacher layers. This confirms that enhanced LoRA has the intended effect.

\noindent
\textbf{Segmentation results using convolutional networks.}   
To showcase the versatility of our approach, we test WeCoLoRA on a medical image segmentation task, by integrating it into the U-Net architecture~\cite{Ronneberger-MICCAI-2015}. The segmentation model employs ResNet-18~\cite{He-CVPR-2016} as backbone. Since the segmentation model is based on convolutional layers, we replace LoRA~\cite{Hu-ICLR-2022} with ConvLoRA~\cite{Aleem-ISBI-2024}. The experiments are performed on the Multiple Sclerosis Lesion Segmentation benchmark (MSLesSeg 2024) \cite{Rondinella-CBM-2023}. We report results in terms of the Dice coefficient in Table~\ref{tab:seg}, where we compare our WeCoLoRA with the strongest baseline, namely WeCo+KD. The results demonstrate that WeCoLoRA obtains higher performance than WeCo+KD, when a convolutional backbone is employed on a segmentation task. This demonstrates the compatibility of WeCoLoRA with both transformer and convolutional architectures, as well as its applicability to diverse tasks, namely classification and segmentation.

\begin{table}[t!]
    \centering
    \begin{tabular}{|l|c|}
    \hline  
         Method & Dice Coefficient \\
        \hline   
        \hline
        WeCo+KD  (ablated) & $0.7665$ \\
        WeCoLoRA (ours) & $\textbf{0.7708}$  \\
        \hline  
    \end{tabular}
         \vspace{-0.2cm}
     \caption{Segmentation results in terms of Dice coefficient obtained with WeCo+KD and WeCoLoRA on the MSLesSeg benchmark. Compression factor: $r\!=\!2$.}
     \label{tab:seg}
        \vspace{-0.6cm}
\end{table}

\begin{table*}[!t]
    \centering
    \begin{tabular}{|l|c|c|c|c|}
    \hline  
         Method & Compression factor $r$ & CIFAR-100 & RESISC-45 & ChestX-ray14 \\
        \hline   
        \hline
        WeCo+KD (ablated) & 6 &$22.3$ & $52.8$ & $62.8$ \\
        WeCoLoRA (ours) & 6 &$24.1$  & $53.6$ & $\textbf{63.7}$ \\
        \hline
        WeCoLoRA+supervised fine-tuning & 6 &$18.4$  & $\textbf{54.5}$ & $63.3$  \\
        WeCoLoRA+classification head transfer & 6 & $\textbf{27.1}$ & $54.2$ & $63.6$ \\
        \hline
        \hline
        WeCo+KD (ablated) & 8 &$20.6$ &$49.2$ &$62.5$ \\
        WeCoLoRA (ours) & 8 &$25.0$  & $\textbf{53.8}$ & $63.1$ \\ 
        \hline
        WeCoLoRA+supervised fine-tuning & 8 &$21.8$ & $53.3$ & $62.4$ \\
        WeCoLoRA+classification head transfer & 8 & $\textbf{25.9}$ & $53.6$ & $\textbf{63.3}$ \\
        \hline  
    \end{tabular}
         \vspace{-0.2cm}
     \caption{Results of various training paradigms in terms of accuracy (in percentages) on CIFAR-100 \cite{Krizhevsky-TR-2009}, NWPU-RESISC45 \cite{Cheng-PIEEE-2017} and in terms of mean AUC (in percentages) on ChestX-ray14 \cite{Wang-CVPR-2017}. Results are reported for the unsupervised ViT-L teacher or backbone ~\cite{Dosovitskiy-ICLR-2021}. During the fine-tuning or distillation procedure, only 1\% of ImageNet-1K \cite{Deng-CVPR-2009} training set is used.}
     \label{tab:vitl}
\end{table*}

\noindent
\textbf{Deeper teacher, higher compression factors.} To demonstrate the applicability of WeCoLoRA to deeper teachers, and its robustness to higher compression factors, we perform additional experiments with the ViT-L teacher based on supervised pre-training, considering compression factors of $r=6$ and $r=8$. In Table \ref{tab:vitl}, we report the results of WeCo+KD and WeCoLoRA on CIFAR-100, RESISC-45 and ChestX-ray14. WeCoLoRA outperforms WeCo+KD for all compression factors, thus showcasing consistent performance gains across various compression factors and teacher models.

\noindent
\textbf{WeCoLoRA based on fine-tuning instead of distillation.} The feature distillation performed by WeCoLoRA is unsupervised, \ie~our framework does not require classification labels during distillation. An alternative approach is to employ supervised fine-tuning instead of unsupervised feature distillation. As shown in Table \ref{tab:vitl}, the fine-tuning combined with WeCoLoRA produces worse results on CIFAR-100, while leading to similar results on RESISC45 and ChestX-ray14. We thus conclude that the supervised fine-tuning is not always beneficial.

\noindent
\textbf{WeCoLoRA with classification head transfer.} One way to potentially boost the performance of WeCoLoRA is to transfer the classification head from the corresponding teacher model, instead of initializing the classification head of the student model from scratch. This idea is explored in Table \ref{tab:vitl} (last row), where it exhibits performance boosts on CIFAR-100. The results on RESISC45 and ChestX-ray14 do not clearly show the benefit of transferring the classification head. 

\section{Limitations}
\vspace{-0.1cm}

The main limitation of our method is its applicability to architectures that use multiple consecutive blocks with the same configuration, \eg~vision transformers and ResNets \cite{He-CVPR-2016}. This restriction is imposed by our weight copying mechanism. Our ablation results indicate that the weight copying step is very useful in the few-shot distillation scenario, as it significantly boosts performance (see Table 1 from the main article). Simply removing the weight copying step is not a viable option, since the performance would drastically degrade. To make our framework applicable to any architecture, the weight copying mechanism could be enhanced with adaptor blocks, which would be able to reshape the copied weights to the appropriate size. However, the adaptor blocks need to be tailored for each specific pair of teacher and student models. This will increase the complexity of the hyperparameter tuning stage, which, in the current form, is quite straightforward, \ie aside from typical hyperparameters, such as the learning rate and the mini-batch size, WeCoLoRA only adds the compression ratio $r$ and the rank of the low-rank matrices $k$ as extra hyperparameters.

\end{document}